\documentclass[runningheads]{llncs}

\usepackage{eccv}

\usepackage{eccvabbrv}

\usepackage{graphicx}
\usepackage{booktabs}

\usepackage[accsupp]{axessibility}  

\usepackage{hyperref}

\usepackage{orcidlink}

\usepackage{xcolor} 

\usepackage{multirow} 
\usepackage{tabularx} 
\usepackage{array}
\usepackage{caption}

\begin{document}

\title{Keep-or-Drop? Adaptive Tokenizer for Compact Video Representation} 


\author{Yeonkyeong Lee \and
Hyunsung Go \and
Jongmin Kim \and
Sewoong Lim \and
Donghoon Lee\thanks{Corresponding author.}}

\authorrunning{Y.~Lee et al.}

\institute{Kakao Corp., Republic of Korea \\
\email{\{mag.ic, cog.map, lukas.ai, amita.lim, dev.e\}@kakaocorp.com} \\
Project page: \url{https://kakao.github.io/KATok/}}

\maketitle

\begin{abstract}
Latent diffusion models have emerged as a dominant framework for high-fidelity image and video synthesis, operating in compact latent spaces with variational autoencoders (VAEs) to enhance computational efficiency without compromising visual quality.
However, conventional VAEs are suboptimal for video data as they employ fixed compression ratios that cannot adapt to the varying complexity of spatio-temporal content. We present KATok (Keep-or-Drop?\ Adaptive Tokenizer for Compact Video Representation), a transformer-based VAE that incorporates an adaptive token selector which is jointly learned with latent tokens. By evaluating each token's content-richness as {\it keep-or-drop probability}, the token selector effectively discards uninformative tokens, naturally allowing data-dependent compression. 
Applying adaptive tokenization to diffusion models may cause spatial misalignment, as token dropping can disturb the original spatio-temporal structure.
To alleviate this issue, we propose two position-prediction strategies: cascaded and joint generation, to ensure spatial consistency.
We empirically show that our model achieves strong reconstruction and generation quality at a state-of-the-art compression ratio.
Further analysis on video data reveals that this improvement is primarily achieved by reducing spatio-temporal redundancy and removing uninformative tokens, as supported by both quantitative and qualitative results.

\end{abstract}

\begin{figure*}[t]
    \centering
    \begin{minipage}[t]{0.64\textwidth}
        \centering
        \includegraphics[width=\textwidth]{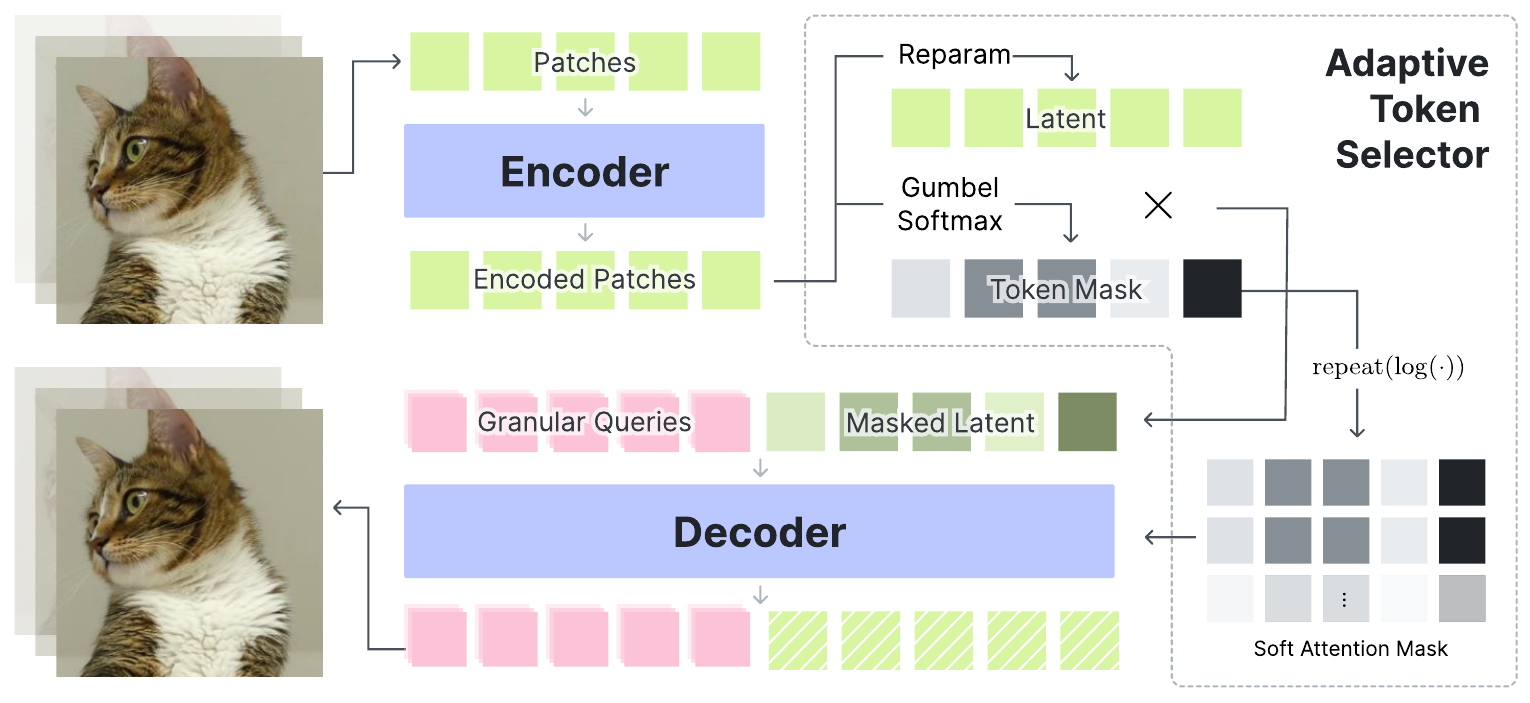}
        \caption{\textbf{Overall architecture of KATok.} A video is divided into spatio-temporal patches and encoded into continuous latent tokens. The adaptive token selector predicts per-token keep probabilities via Gumbel–Softmax relaxation, producing a soft token-drop mask shared with the decoder attention for consistent token selection. The decoder reconstructs the input via learnable query tokens by attending to the masked latent tokens, forming an end-to-end differentiable pipeline for adaptive token selection.}
        \label{fig:overview}
    \end{minipage}
    \hfill
    \begin{minipage}[t]{0.33\textwidth}
        \centering
        \includegraphics[width=\textwidth]{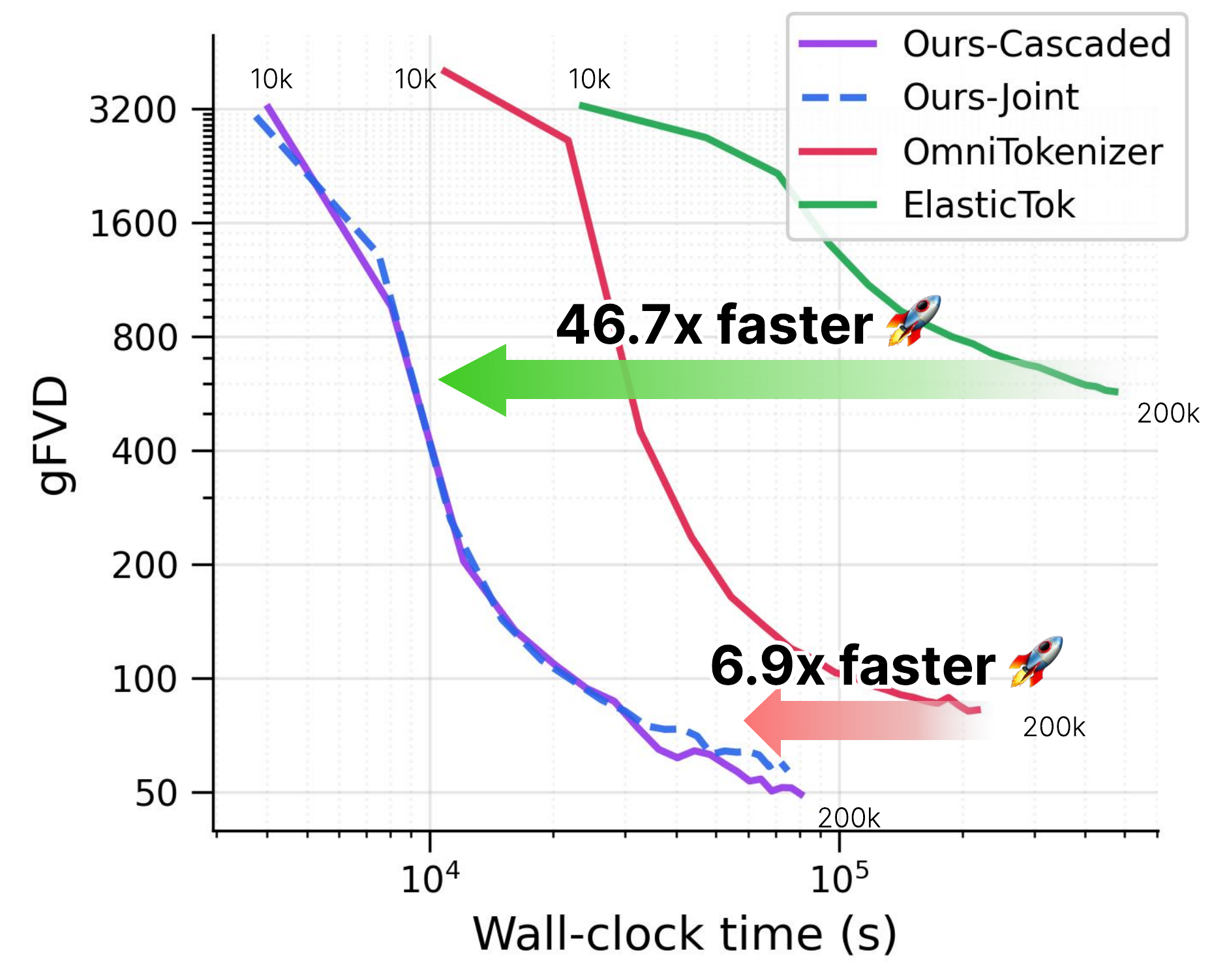}
        \caption{\textbf{Generation quality over training time.} KATok converges faster and achieves superior final generation quality than both the dense baseline OmniTokenizer and the flexible baseline ElasticTok. The x-axis reports wall-clock time measured on 8 H200 GPUs.}
        \label{fig:wall_clock}
    \end{minipage}
\end{figure*}

\section{Introduction}
\label{sec:intro}

Latent diffusion models (LDMs)~\cite{rombach2022ldm, saharia2022imagen, esser2024sd3, batifol2025flux} have achieved remarkable progress in high-fidelity image and video synthesis by performing diffusion in latent spaces with variational autoencoders (VAEs)~\cite{kingma2013vae, esser2021vqgan}. Compared to early pixel-space diffusion approaches~\cite{ho2020ddpm, song2020ddim} that were limited by the high dimensionality of visual data, conventional VAE architectures in LDMs~\cite{esser2021vqgan, rombach2022ldm} reduce the spatial and temporal dimensions of input videos. 
This latent compression drastically decreases memory and computation requirements while maintaining visual fidelity, making LDMs an effective foundation for scalable video generation.

While VAEs help reduce the computational cost substantially by operating in a latent space \cite{chen2024dcae, hacohen2024ltx, yu2024titok}, this approach remains limited for video data---particularly high-resolution or long sequences---since the number of tokens still scales with spatial and temporal dimensions. Given the strong spatio-temporal redundancy in videos, this non-adaptive latent representation inevitably allocates capacity to uninformative regions, underscoring the need for content-adaptive representations.

Recent studies have attempted to address this limitation by introducing variable-length tokenization~\cite{bachmann2025flextok, miwa2025onedpiece, yan2024elastictok}. These approaches provide flexibility by allowing users to determine the number of tokens based on the desired reconstruction quality—retaining more tokens for higher fidelity and fewer tokens for coarser results. However, such \emph{flexibility} does not imply \emph{adaptivity}: while they provide user-controlled compression, determining the optimal token count for each sample often requires additional model or inference-time search, introducing overhead in downstream tasks. We consider \emph{adaptive tokenization} a paradigm where the model automatically determines how many tokens are needed for each sample based on its content complexity. 
It selectively removes redundant or uninformative tokens to produce compact yet expressive representations.

Building on this concept, we propose KATok, an end-to-end adaptive tokenization method designed to operate on continuous latent representations compatible with diffusion models, as illustrated in Fig.~\ref{fig:overview}.
KATok employs an Adaptive Token Selector, which learns to predict token importance directly from each sample and determines which tokens to retain or discard through a differentiable gating mechanism based on the Gumbel–Softmax \cite{jang2016categoricalgumbel} relaxation.
A sparsity regularization further encourages compact yet informative representations, allowing adaptive compression to emerge naturally during training.

We employ KATok to construct the latent space for diffusion-based video generation.
However, naively applying adaptive tokenization to the generative model can cause spatial misalignment: under sparse tokenization, the model may fail to recover the original token positions, leading to content--position mismatch.
To address this, we propose two remedies: (i) joint prediction of content and positions, and (ii) a cascaded generation scheme that separates position selection from content generation.
Both strategies lead to substantial gains in spatial coherence and overall sample quality, especially under strong sparsification.
Our model enables higher-quality video generation while using substantially fewer tokens than conventional tokenizers.

Our main contributions are summarized as follows:
\begin{itemize}
    \item \textbf{Adaptive VAE for token-wise compression:} We propose an adaptive VAE that learns token importance through soft attention gating and sparsity regularization, enabling sample-adaptive compression without predefined token budgets.

\item \textbf{Mitigating content--position misalignment under sparse tokenization:} We address spatial misalignment caused by adaptive sparsification by proposing (i) joint content--position prediction and (ii) a cascaded scheme that decouples position selection from content generation.

    \item \textbf{Efficient training and generation:} Our approach enables high-quality video generation while using substantially fewer tokens, achieving \textbf{6.9$\times$} faster training and \textbf{3.2$\times$} faster inference than the strong and widely adopted trans-former-based tokenizer \cite{wang2024omnitokenizer} (see Fig.~\ref{fig:wall_clock}). 
\end{itemize}

\section{Related Works}
\label{sec:related}
In this section, we review prior work in three related areas: visual tokenizers that enable compact latent modeling, flexible and adaptive tokenization approaches that allocate representational capacity dynamically, and token dropping or merging methods that improve computational efficiency.
\subsection{Visual Tokenizers}
A key driver of progress in modern generative models \cite{rombach2022ldm, wan2025wan, batifol2025flux, hong2022cogvideo, saharia2022imagen, lipman2022flow, esser2024sd3} has been the use of autoencoder-based tokenizers \cite{kingma2013vae, li2025wfvae}, which compress high-dimensional inputs into compact latent spaces that enable large-scale training.  
Latent Diffusion Models (LDMs) \cite{rombach2022ldm} showed that convolutional VAEs can serve as effective visual tokenizers, enabling high-resolution synthesis with manageable compute.  
Since then, VAE-based tokenizers have become the standard in image and video generation, balancing compression efficiency with reconstruction fidelity.

Recent extensions such as DC-AE \cite{chen2024dcae} and LTX-Video \cite{hacohen2024ltx} improved compression and temporal modeling but still scale poorly with spatial resolution and sequence length, leaving cross-frame redundancy underutilized.  
To address this limitation, recent works have explored flexible tokenization strategies that allocate representational capacity adaptively across space and time.  
We discuss these approaches in the following subsection.

\subsection{Flexible and Adaptive Tokenization}
Early work on efficient visual representation reformulated images and videos as one-dimensional token sequences to improve compactness and scalability \cite{yu2024titok, he2025flowtok}.  
Building on this architecture, recent studies introduced flexible-length tokenization, where the number of retained tokens varies with content importance.  
Methods such as One-D-Piece \cite{miwa2025onedpiece}, FlexTok \cite{bachmann2025flextok}, SEED \cite{ge2023seed}, and ALIT \cite{duggal2025adaptive} employ dropout-based, causal, or iterative allocation strategies to prioritize informative tokens and enable variable-length reconstructions.  
ElasticTok \cite{yan2024elastictok} further extends this concept to video by combining tail-drop truncation with causal attention.

Despite these advances, flexible-length tokenizers still rely on predefined budgets or inference-time search.
KATok instead performs continuous and fully differentiable token selection within a transformer encoder, achieving adaptive compression while preserving fine-grained spatial and temporal structure.

\subsection{Token Dropping and Merging}
A separate line of works focus on token pruning and merging techniques, such as DynamicViT \cite{rao2021dynamicvit}, EViT \cite{liang2022evit}, and ToMe \cite{bolya2022tome}, which accelerate transformers by reducing redundancy at inference time.  
Unlike these inference-only approaches, KATok learns sparsity during training, yielding compact latent representations that remain efficient at both training and inference.

\section{Adaptive Tokenizer for Video Representation}
\label{sec:method}

We aim to develop an adaptive tokenization framework that dynamically adjusts the number of tokens according to content complexity, eliminating the need for preset budgets or inference-time search. Our goal is to achieve compact yet expressive latent representations that maintain spatio-temporal consistency while improving efficiency in video generation tasks.
An overview of the pipeline appears in Fig.~\ref{fig:overview}.

\textbf{Problem setup.} Let a video $X \! \in \! \mathbb{R}^{T\times H\times W\times C}$ be patch-embedded into $\{x_i\}_{i=1}^{N}$. In contrast to conventional tokenizers which produce a fixed-length latent $Z \! \in \! \mathbb{R}^{N\times d_\text{latent}}$, 

we learn a content-adaptive token count $N_{\mathrm{eff}} \! \le \! N$, where $N_{\mathrm{eff}}(X) \! := \! \sum_{i=1}^{N} m_i(X)$ is defined by the per-token masks $m_i(X) \in \{0, 1\}$, with the objective of minimizing $N_{\mathrm{eff}}(X)$ while maintaining fidelity.

\subsection{Adaptive Keep-or-Drop Tokenization}
\label{sec:adaptive_tokenization}
Given patch embeddings $\{x_i\}_{i=1}^{N}$, the encoder $E_\theta$ produces intermediate embeddings $e_i \in \mathbb{R}^{d_\text{model}}$, which are projected to the parameters of a diagonal Gaussian,
$\mu_i, \sigma_i \in \mathbb{R}^{d_\text{latent}}$, via $(\mu_i, \sigma_i) = f_\theta(e_i)$.
A lightweight token-importance network $g_\theta$ predicts keep/drop logits per token,
$\alpha_i = g_\theta(e_i) \in \mathbb{R}^2$.
During training, we obtain differentiable soft masks $\widetilde{m}_i$ using the Gumbel-Softmax relaxation~\cite{jang2016categoricalgumbel}:
\begin{equation}
[\widetilde{m}_i, 1 - \widetilde{m}_i] = \mathrm{GumbelSoftmax}(\alpha_i;\tau).
\end{equation}
We sample continuous latents via reparameterization,
$\hat{z}_i = \mu_i + \sigma_i \odot \epsilon_i,\; \epsilon_i \sim \mathcal{N}(0,I)$,
and apply soft gating as $z_i = \widetilde{m}_i \cdot \hat{z}_i$.
At inference time, we discard tokens using hard masks
$m_i = \mathbb{I}(\alpha_{i0} \geq \alpha_{i1})$.

\textbf{Soft attention masking.}
To ensure that token dropping consistently affects decoding, we use the same soft masks to modulate the decoder attention logits.
Specifically, attention scores $A_{ij}$ are shifted as
$A_{ij} \leftarrow A_{ij} + b_j$ where $b_j = \log(\widetilde{m}_j + \varepsilon)$ with a small $\varepsilon>0$,
so that $\widetilde{m}_j$ acts as a differentiable \emph{soft attention mask}. 
Empirically, removing this masking leads to training instability and severe reconstruction degradation (Table~\ref{tab:ablation}).

\textbf{Sparsity regularization.}
Our goal is to minimize the expected number of active tokens while letting the model decide which tokens to keep.
We apply an $\ell_1$ penalty to the soft masks:
\begin{equation}
\mathcal{L}_\text{sparse}
= \mathbb{E}_X \Big[ \sum_{i=1}^N \widetilde{m}_i(X) \Big]
\approx \mathbb{E}_X \big[ N_\mathrm{eff}(X) \big].
\end{equation}
The corresponding weight $\lambda_{\text{sparse}}$ is gradually annealed during training:
we start with a weak penalty (allowing nearly all tokens to be kept for $\sim$5k iterations) and then progressively increase it to enforce sparsity once optimization stabilizes.

This design enables end-to-end, learnable token selection, making the number of active tokens an outcome of training rather than a predefined budget, and eliminating inference-time search or handcrafted dropout schedules.

\subsection{Training Objective}
\label{sec:objective}
We train the tokenizer--VAE with a weighted combination of reconstruction, KL divergence, adversarial, sparsity, and representation losses:
\begin{equation}
\mathcal{L}
= \mathcal{L}_{\text{recon}}
+ \lambda_{\text{KL}}\mathcal{L}_{\text{KL}}
+ \lambda_{\text{sparse}}\mathcal{L}_{\text{sparse}}
+ \lambda_{\text{adv}}\mathcal{L}_{\text{adv}}
+ \lambda_{\text{align}}\mathcal{L}_{\text{align}} .
\end{equation}
Here, $\mathcal{L}_{\text{recon}}$ comprises pixel- and perceptual-level losses ($\ell_1$ and perceptual similarity),
$\mathcal{L}_{\text{KL}}$ regularizes the (masked) latent distribution,
$\mathcal{L}_{\text{sparse}}$ promotes token efficiency (Section~\ref{sec:adaptive_tokenization}),
and $\mathcal{L}_{\text{adv}}$ encourages sharp reconstructions.

\textbf{Video-aware perceptual loss.}
We replace LPIPS~\cite{zhang2018lpips} with Video-LPIPS, which computes perceptual similarity using spatio-temporal features extracted from an S3D network~\cite{xie2018rethinkings3d} to better capture temporal coherence and motion consistency across frames.

\textbf{Latent regularization.}
To stabilize training and improve generative fidelity, we regularize latent decoding with (i) latent noise augmentation and (ii) a representation alignment loss.
Following LTX-Video~\cite{hacohen2024ltx}, we inject small Gaussian noise into latent tokens during training, with a noise level uniformly sampled from $[0, 0.2]$.
In addition, we consider a representation alignment loss \cite{yao2025reconstruction}, denoted as $\mathcal{L}_\text{align}$, using features from a \emph{frozen} vJEPA-2~\cite{assran2025v} encoder. In practice, these regularizers slightly degrade reconstruction metrics, but consistently improve generation quality and accelerate convergence (see Table \ref{tab:gen_ablation_ucf}); we therefore adopt them as a deliberate trade-off to better optimize for generative fidelity.

\subsection{Encoder-Decoder Architecture}
The overall architecture is illustrated in Fig.~\ref{fig:overview}.
The encoder is a self-attention transformer equipped with 3D rotary positional embeddings (RoPE) \cite{su2024rope} and two register tokens \cite{darcet2023vision} acting as global reference points. Our encoder captures spatio–temporal redundancy across frames and outputs continuous latent tokens sampled via the reparameterization trick.
After token selection, the decoder reconstructs inputs through double-stream blocks, adopted from FLUX~\cite{flux2024}, in which repeated learnable query tokens and the latent tokens flow through their respective towers. The latent tokens omit positional encoding in the decoder, while the learnable query tokens employ 3D RoPE as in the encoder.

Patchification and unpatchification are implemented as linear projections, ensuring full differentiability throughout the pipeline.
Table \ref{tab:ablation} shows that removing the register tokens increases token usage and slightly degrades spatial coherence, confirming that they act as global anchors that stabilize structure under aggressive token reduction.

\textbf{Asymmetric coarse-to-fine decoding.}
We introduce an \emph{asymmetric} patching strategy that keeps encoding compact while improving decoding detail.
The encoder tokenizes videos into coarse 3D patches of size $16^2 \times 8$ and produces continuous latent tokens, which are sparsified by our keep-or-drop tokenizer.
The decoder reconstructs on a finer grid of size $8^2\times 4$ by increasing only the number of learnable query tokens $\{q_j\}_{j=1}^{M}$, where $M$ matches the fine-grid patch count.
This preserves a compact latent set for efficiency yet enables fine-grained reconstruction, improving both reconstruction and generation quality.

\begin{figure*}[t]
    \centering
    \resizebox{\textwidth}{!}{%
    \begin{subfigure}[t]{0.5\textwidth}
        \centering
        \includegraphics[width=\textwidth]{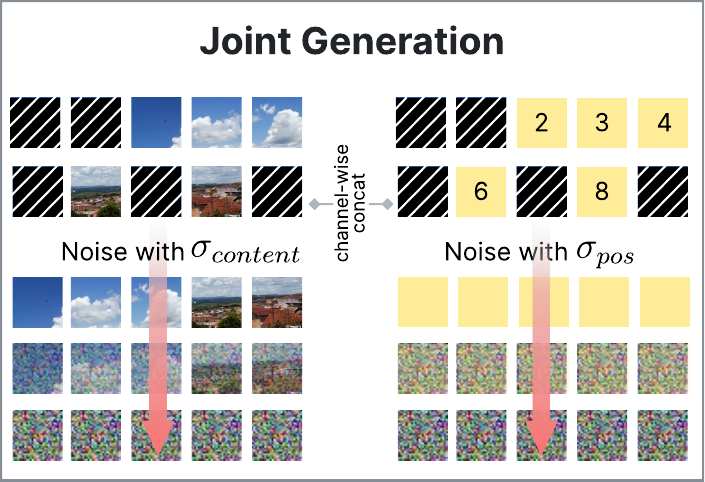}
        \caption{Diffusion training with joint generation.}
        \label{fig:gen-overview-1}
    \end{subfigure}
    \hfill
    \begin{subfigure}[t]{0.5\textwidth}
        \centering
        \includegraphics[width=\textwidth]{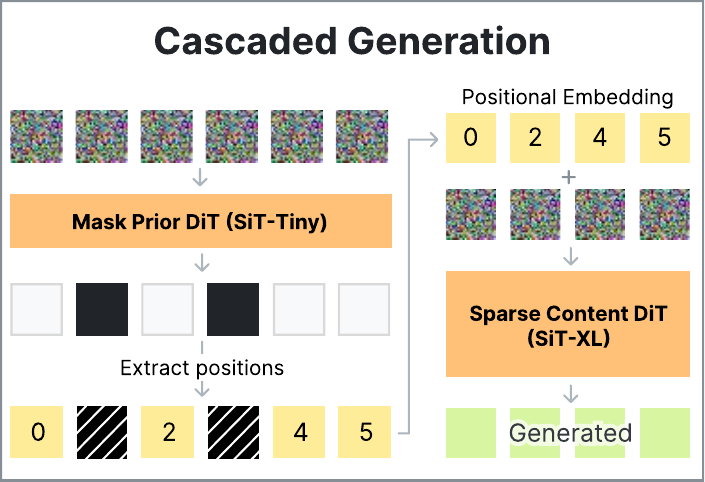}
        \caption{Diffusion inference with mask prior.}
        \label{fig:gen-overview-2}
    \end{subfigure}
    } 
    \caption{\textbf{Two variants of sparse content generation.}
\textbf{(a) Diffusion training with joint content–position generation.} During training, content and position tokens are modeled jointly but follow decoupled noise schedules—$\sigma_{\text{content}}$ for latent denoising and $\sigma_{\text{pos}}$ for spatio-temporal coordinate prediction—stabilizing training and improving spatial alignment.
\textbf{(b) Cascaded generation of mask and content.} While ground-truth token masks and positions are used during training, at inference the mask prior model generates a plausible token mask, which is then thresholded to extract positions for positional encoding in the sparse content generation model.}
    \label{fig:gen-overview}
\end{figure*}

\section{Diffusion with Sparse Latent Tokens}
\label{sec:gen}
We employ the learned VAE as the tokenizer for training downstream generative models, specifically using the \emph{flow matching} framework~\cite{lipman2022flow}.
Given latent tokens $z$, flow matching learns a continuous-time velocity field $v_\theta(z_t, t)$ that transports Gaussian noise to clean latents.
Let $z_0 \in \tilde{Z}$ denote clean latent tokens and $z_1 \sim \mathcal{N}(0,I)$ a noise sample.
A linear interpolation defines intermediate states as $z_t = (1-t) z_0 + t z_1,\; t \in [0,1]$.
The training objective minimizes the discrepancy between the model prediction and the true velocity:
\begin{equation}
\mathcal{L}_{\text{content}} = \mathbb{E}_{z_0, z_1, t}\Big[ \big\| v_\phi(z_t, t) - (z_1 - z_0) \big\|^2 \Big].
\label{eq:flowmatching}
\end{equation}
At inference, the learned vector field is integrated to generate latents from Gaussian noise.

Because adaptive tokenization selectively removes redundant tokens, directly training a generative model on the resulting sparse latents can induce \emph{content--position misalignment} under aggressive sparsification (Fig. \ref{fig:misalign}): the remaining tokens may no longer align with their original spatial locations, leading to unstable boundaries and temporal inconsistencies.

\subsection{Joint content--position generation with timestep decoupling.}
As a first approach, we jointly predict latent contents and their original positions during flow-matching training.
The position co-objective $\mathcal{L}_{\text{pos}}$ mirrors Eq.~\ref{eq:flowmatching} but replaces the latent targets with ground-truth positions $\rho \in \mathbb{R}^{N \times 3}$:
\begin{equation}
\mathcal{L}_{\text{pos}} =
\mathbb{E}_{\rho_0, \rho_1, t}\Big[
\big\| v_\phi(\rho_t, t) - (\rho_1 - \rho_0) \big\|^2
\Big],\quad
\rho_t = (1-t)\rho_0 + t\rho_1,
\end{equation}
and we optimize $\mathcal{L}_\text{flow}=\mathcal{L}_\text{content}+\lambda_\text{pos}\mathcal{L}_\text{pos}$.
We further apply \emph{timestep decoupling} (Fig.~\ref{fig:gen-overview-1}), assigning separate noise schedulers for content and position to decouple the denoising dynamics and allow differing noise-level evolutions at inference.
While this joint objective improves spatial consistency (Fig.~\ref{fig:misalign}), its performance is sensitive to the choice of scheduler hyperparameters and the optimal combination of the two noise schedules, which typically requires careful tuning.

\subsection{Cascaded mask-prior conditioning.}
\begin{figure*}[t]
    \centering
    \includegraphics[width=\textwidth]{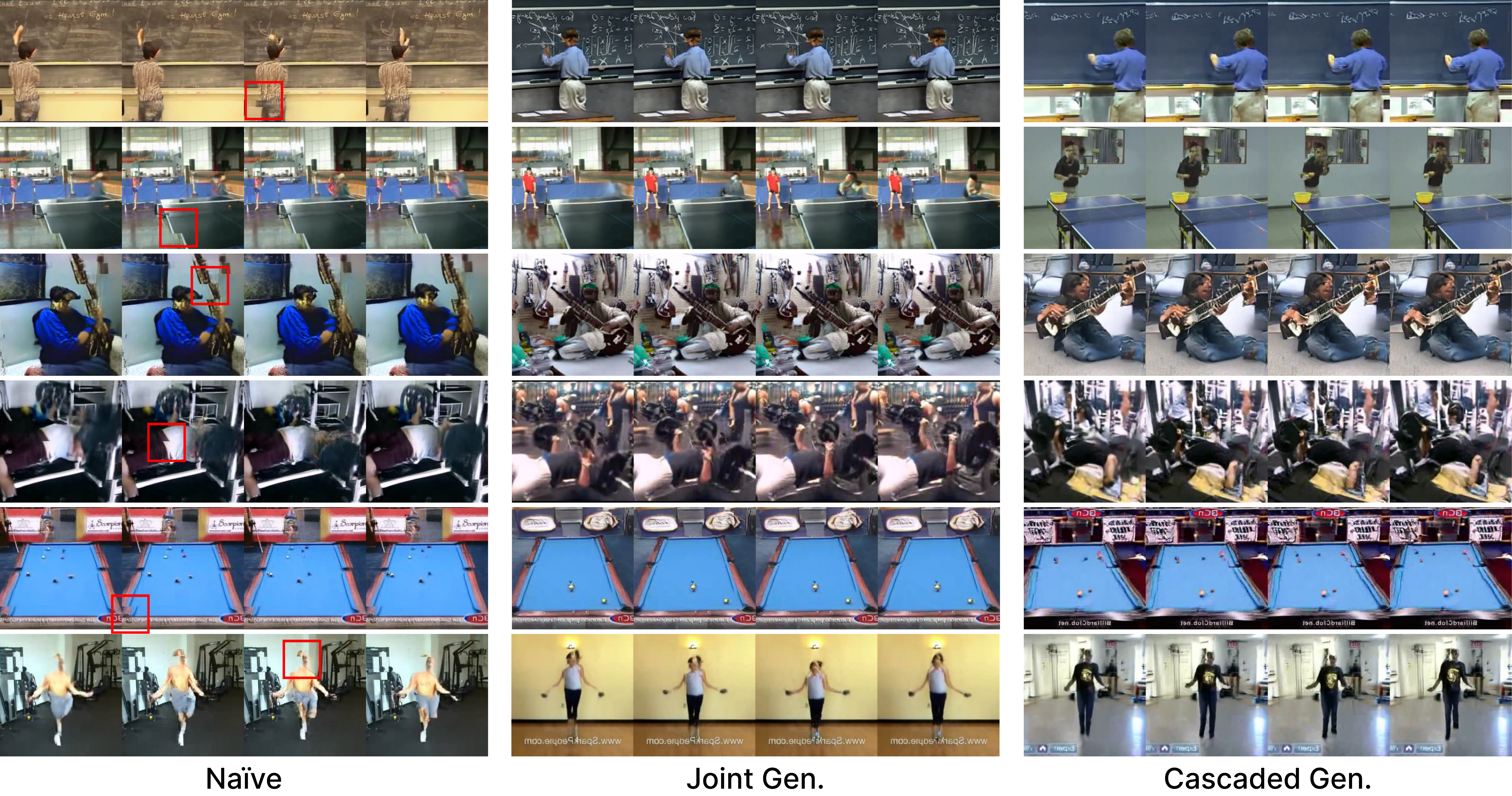}
    \caption{\textbf{Effect of joint content--position and cascaded generation.} The naive model for generating sparse latents often leads to content--position misalignment, as highlighted by the red boxes, while the proposed joint content--position and cascaded generation strategies mitigate this issue and improve overall generation quality.}
    \label{fig:misalign}
\end{figure*}

To avoid hyperparameter search over content- and position-noise scheduling, we propose a cascaded alternative (Fig.~\ref{fig:gen-overview-2}). 
A lightweight (8.3M) \emph{mask prior} model first predicts a binary selection mask over the token grid, indicating which spatial--temporal positions should be active.
We then select the corresponding set of positions and inject it into the main flow model as explicit positional conditioning.
During training, we condition the main flow model on positional embeddings computed from ground-truth token positions, while at inference we use positional embeddings computed from positions predicted by the mask prior.
By decoupling \emph{position selection} from \emph{content generation}, this cascade provides a stable structural prior and substantially reduces content--position misalignment under heavy sparsification (Fig.~\ref{fig:misalign}).
In our experiments, mask-prior conditioning is both more robust and achieves higher generative fidelity, so we adopt it as our default generation strategy, while keeping joint position prediction as a strong ablation.

\section{Experiments}
\label{sec:exp}

Our model is trained on the Panda-70M dataset \cite{chen2024panda70m} using a patch size of $16^2\times8$. The encoder--decoder network scales follow the ViT-B setting~\cite{liang2022evit}. Although the model supports end-to-end training, we adopt a three-stage training strategy for computational efficiency. The first stage uses single-resolution of $256^2\times16$, the second stage extends training to multi-resolution, and the final stage performs GAN-based fine-tuning for perceptual quality. For generative modeling, we use KATok as a tokenizer to train diffusion-based models on the SkyTimelapse \cite{dtvnetsky}, UCF-101 \cite{soomro2012ucf101}, and Kinetics-600 \cite{kay2017kinetics} datasets. Complete implementation details and hyperparameters are provided in the supplementary material.

\begin{figure*}[t]
    \centering
    \begin{minipage}[t]{0.4\textwidth}
        \centering
        \includegraphics[width=\textwidth]{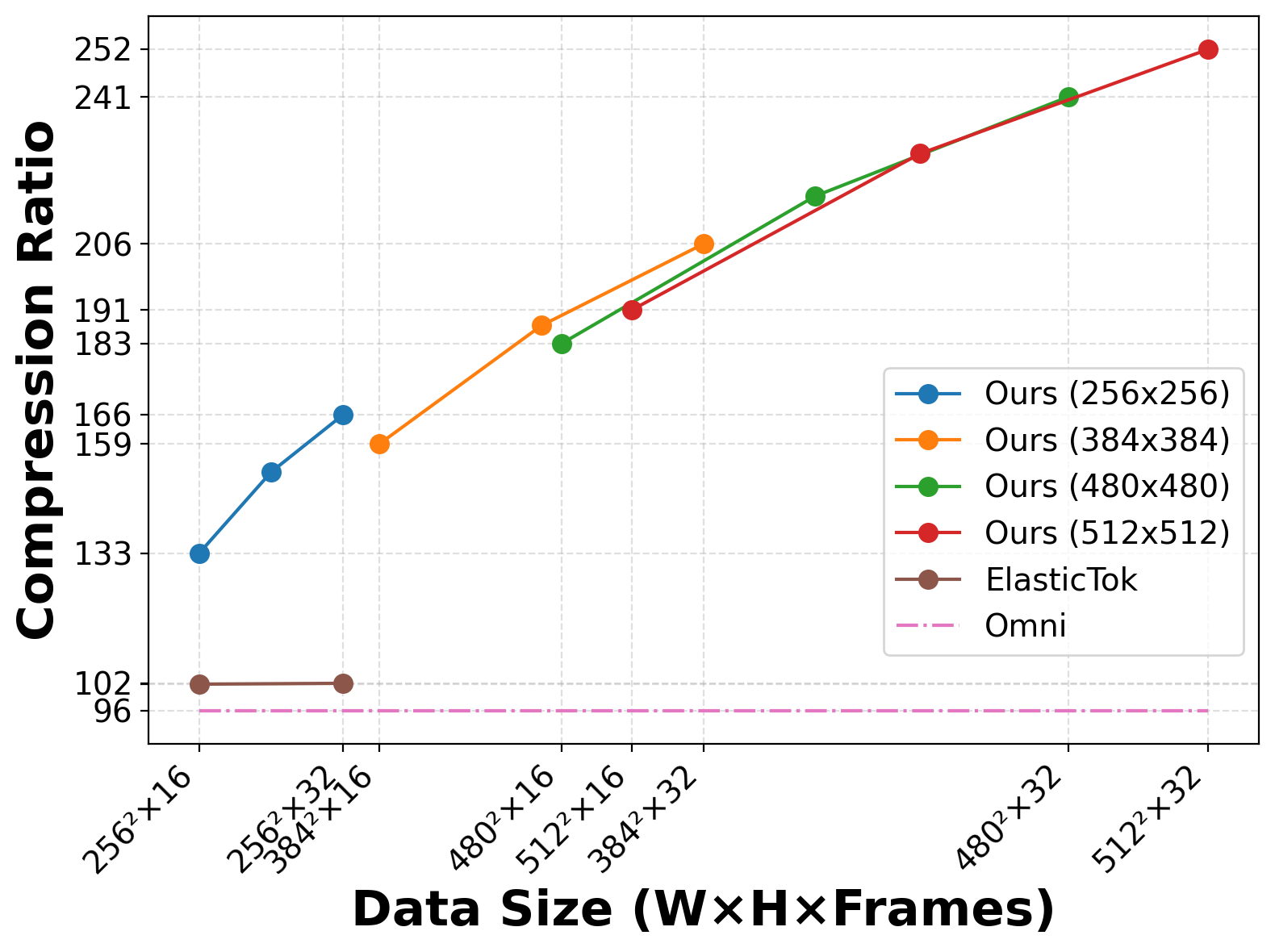}
        \caption{\textbf{Token scaling with video size.}
As spatial and temporal resolution increase, our method achieves progressively higher compression ratios, demonstrating efficient scalability across input sizes. In contrast, OmniTokenizer maintains a fixed compression ratio, and ElasticTok remains nearly flat. (Conventions and compression ratio definition follow Table~\ref{tab:reconstruction_video}.)}
        \label{fig:CR}
    \end{minipage}
    \hspace{0.01\textwidth}
    \begin{minipage}[t]{0.55\textwidth}
        \centering
        \includegraphics[width=\textwidth]{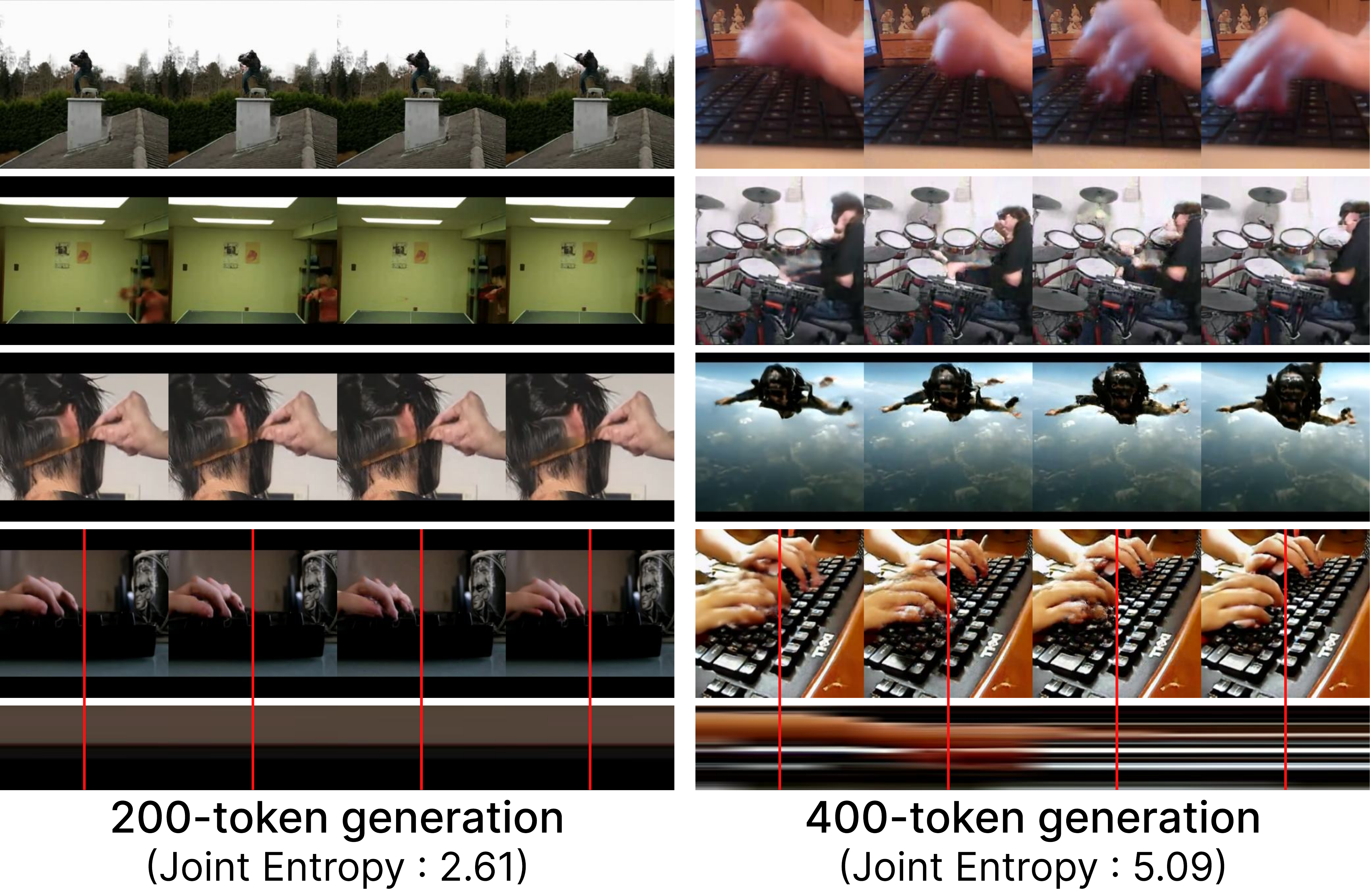}
        \caption{\textbf{Emergent controllability in video generation (UCF101).}
        Increasing the number of tokens at generation time modulates motion and visual detail without retraining or extra conditioning: 200-token samples yield simpler, low-motion videos, whereas 400-token samples produce more dynamic and visually rich sequences. Bottom rows show spatio-temporal slices along y–t axes taken along the red lines, revealing stronger temporal variation with higher token counts.}
        \label{fig:generation-ntokens}
    \end{minipage}
\end{figure*}

\begin{figure*}[t]
    \centering
    \includegraphics[width=\textwidth]{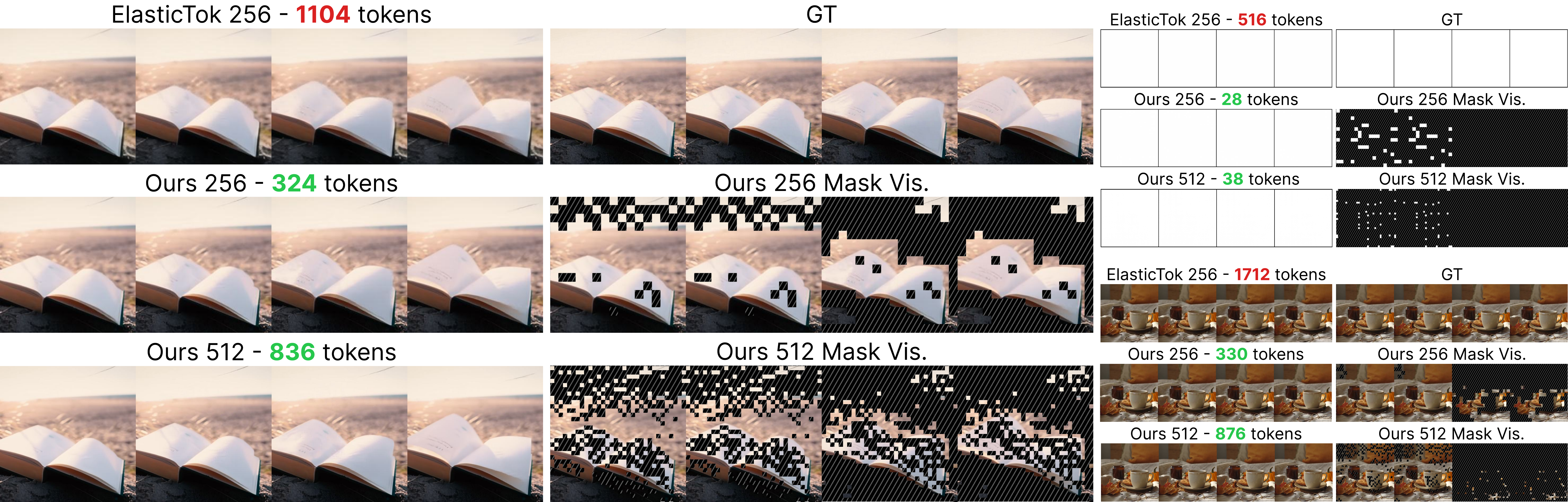}
    \caption{
\textbf{Qualitative reconstruction results.}
Our results are shown at $256^2\times16$ and $512^2\times16$, while ElasticTok is at $256^2\times16$. The learned token mask $m_i$ is visualized by mapping token $i$ back to its spatial patch; black tiles denote suppressed tokens. Our tokenizer attains better perceptual quality with far fewer tokens (e.g., 1104$\rightarrow$836 on the ocean clip; 1712$\rightarrow$330 indoors); the top-right example is a solid-white video, an extreme case where our method needs only 28 tokens.
}
    \label{fig:recon}
\end{figure*}

\begin{figure*}[t]
    \centering
    \includegraphics[width=0.9\textwidth]{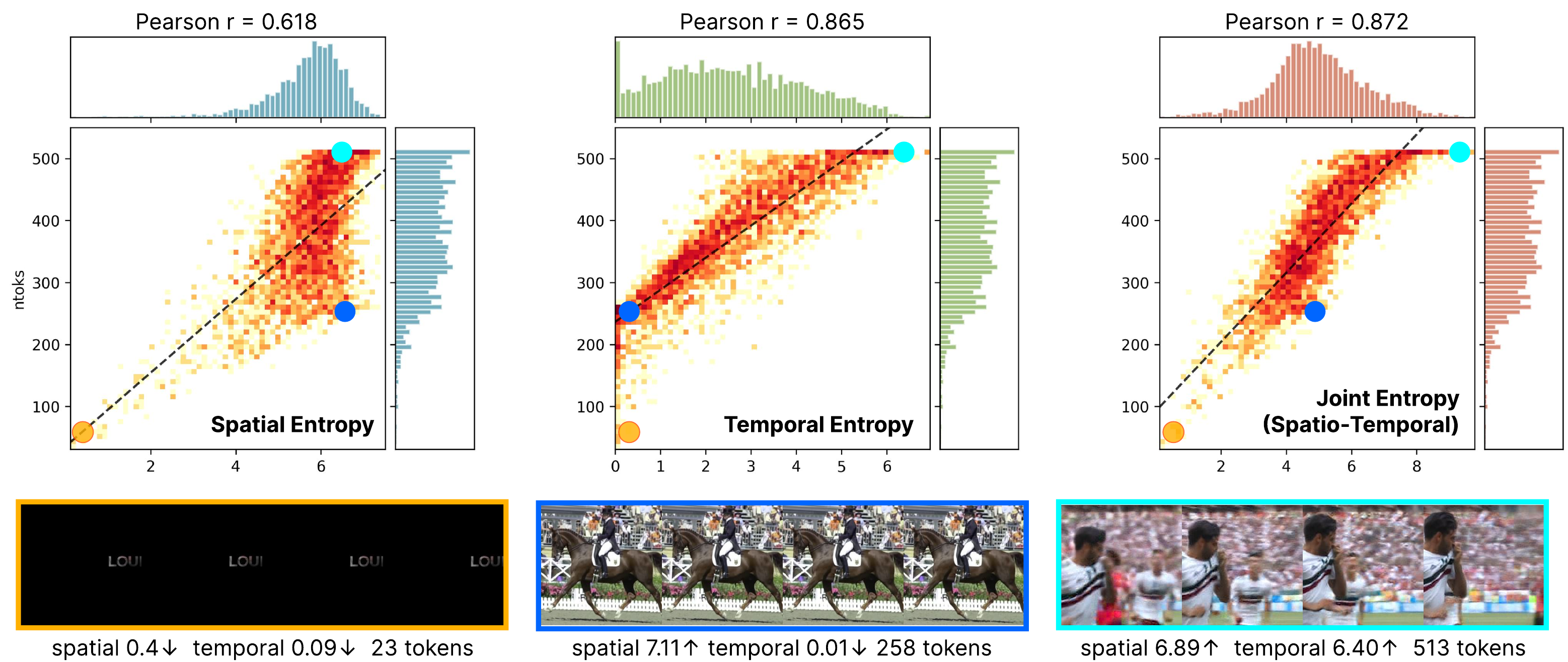}
    \caption{\textbf{Token count vs. content entropy.}
Scatter plots show the relationship between spatial, temporal, and spatio–temporal entropy (x-axis) and the number of tokens used by our tokenizer (y-axis).
Each panel reports Pearson correlations.
Highlighted examples illustrate the trend: \textcolor{orange}{orange} — simple, low-motion video uses few tokens; \textcolor{cyan}{cyan} — complex scene with strong motion uses many tokens; \textcolor{blue}{blue} — complex but mostly static scene uses fewer tokens despite high spatial entropy.
Overall, token count correlates much more strongly with \emph{temporal} entropy than with spatial entropy (pearson $r\!\approx\!0.87$ vs.\ $0.62$).}
    \label{fig:entropy}
\end{figure*}

\subsection{Adaptive Compression and Reconstruction Analysis}

\begin{table*}[t]
\centering
\footnotesize
\caption{\textbf{Reconstruction on the \texttt{panda70m} validation set.} \#Tokens denotes the token count; for Ours and ElasticTok-KL, it indicates the average number of tokens used. \texttt{Comp.$\uparrow$} is the compression ratio,
$\mathrm{Comp.}=\frac{H\times W\times T\times 3}{(\#\text{tokens})\times(\text{token channels})}$
where “token channels’’ equals the \texttt{Channels} column. Our tokenizer achieves better reconstruction quality with substantially fewer tokens. \emph{Note:} ElasticTok is available only at $256^2$ spatial resolution, thus results at $512^2\times32$ are not supported (cells marked “–”). \textbf{*}Metrics are evaluated on the first 16/32 frames for fair comparison.}
\label{tab:reconstruction_video}
\begin{tabular}{l|c|c|c|c|c|c|c|c|c|c}
\toprule
Method & Resolution & \#Tokens & Channels & Comp.$\uparrow$ & PSNR$\uparrow$ & LPIPS$\downarrow$ & SSIM$\uparrow$ & rFVD$\downarrow$ \\
\midrule
\multirow{2}{*}{Omni\text{-}VAE}
  & \hspace{0.3em}$256^2\times 17\textsuperscript{*}$ & 5120 & 8 & 96 & 28.10 & 0.05 & 0.88 & 7.84 \\
  & \hspace{0.3em}$512^2\times33$\textsuperscript{*} & 36864 & 8 & 96 & 24.07 & 0.06 & 0.80 & 16.85 \\
\midrule
\multirow{2}{*}{Elastic\text{-}KL}
  &     $256^2\times16$ & 3845.56 & 8 & 102.25 & 30.52 & 0.06 & 0.91 & 12.37 \\
  & $512^2\times32$ & -- & -- & -- & -- & -- & -- & -- \\
\midrule

\multirow{2}{*}{Ours}
  & $256^2\times16$ & \textbf{366.24} & 64 & \textbf{134.21} & \textbf{31.24} & \textbf{0.04} & \textbf{0.94} & \textbf{5.12} \\
  & $512^2\times32$ & \textbf{1554.24} & 64 & \textbf{253.00} & \textbf{33.23} & \textbf{0.05} & \textbf{0.95} & \textbf{6.40} \\
\bottomrule
\end{tabular}
\end{table*}

We evaluate our model on approximately 5,000 videos from the Panda-70M validation split using PSNR, SSIM, LPIPS, and rFVD. 
To our knowledge, ElasticTok \cite{yan2024elastictok} is the only publicly available method that performs variable-length tokenization on video data.
Other flexible tokenization studies \cite{bachmann2025flextok, miwa2025onedpiece, duggal2025adaptive} are designed for images, thus cannot be directly applied to spatio-temporal data.
Therefore, we adopt ElasticTok as the primary comparison baseline for video experiments, using its publicly released KL-based model with binary search for evaluation.

In addition, we include OmniTokenizer-VAE \cite{wang2024omnitokenizer} as a transformer-based fixed-length tokenizer baseline for comparison.
Although OmniTokenizer does not employ adaptive tokenization, it provides a fixed-length transformer baseline under similar architectural constraints, enabling a fair assessment of how KATok’s adaptive mechanism impacts efficiency and reconstruction fidelity.
To keep the comparison controlled, the main paper focuses on baselines that share KATok's modeling paradigm—transformer-based, continuous-token (KL) VAEs. Comparisons against other paradigms, including CNN-based VAEs (Cosmos~\cite{agarwal2025cosmos}, LTX-Video~\cite{hacohen2024ltx}) and VQ-based adaptive tokenizers (EVATok~\cite{xiong2026evatok}, AdapTok~\cite{li2025adaptok}), are provided in the supplementary material.

Table~\ref{tab:reconstruction_video} summarizes reconstruction results on the Panda-70M validation set across spatio--temporal resolutions.
Our method achieves the best reconstruction quality in terms of PSNR and rFVD while using far fewer tokens than the baselines.
At $256^2 \times 16$, Ours reaches 31.24 PSNR with rFVD 5.12, using 366 tokens compared to 5,120 (OmniTokenizer-VAE) and 3,846 (ElasticTok-KL).
At $512^2 \times 32$, Ours remains substantially better than OmniTokenizer-VAE (33.23 vs.\ 24.07 in PSNR; 6.40 vs.\ 16.85 in rFVD), despite using 1,554 tokens versus 32,768.
As resolution increases, the compression ratio improves from 134.21 to 253.00, indicating that adaptive tokenization removes spatio--temporal redundancy more effectively at larger video scales, consistent with Fig.~\ref{fig:CR}.

Fig.~\ref{fig:recon} illustrates how our adaptive token selector behaves across different scenes and resolutions.
The model selectively retains tokens in motion-rich regions while discarding those from static or homogeneous areas, effectively concentrating representational capacity where dynamics occur.
At higher spatial resolution, token dropping becomes stronger even within the same scene—for instance, in the book clip, the first patch that remains active at $256^2$ is further suppressed at $512^2$.
Remarkably, our tokenizer operating at $512^2$ uses only 836 tokens—fewer than ElasticTok’s 1,104 tokens at $256^2$—while achieving better reconstruction quality.
In extremely static cases (e.g., a uniform white video), the model can reconstruct an entire $256^2\times16$ clip with as few as 28 tokens, demonstrating strong spatial–temporal adaptivity and compression efficiency.

To study what governs token allocation, we correlate the expected token count $N_{\mathrm{eff}}(X)\!=\!\sum_i m_i(X)$ with Shannon-entropy measures of clip complexity: spatial complexity via Sobel edge magnitude, temporal complexity via inter-frame intensity change $\Delta I$, and a joint spatio–temporal measure. Across the validation set, $N_{\mathrm{eff}}$ correlates most strongly with temporal entropy (r{=}0.865), more moderately with spatial entropy (r{=}0.618), and closely with spatio–temporal entropy (r{=}0.877) (see Fig.~\ref{fig:entropy}).

\subsection{Video Generation with Adaptive Tokenization}
\label{sec:generation}

\textbf{Setup.}
We evaluate downstream generation using flow matching with a SiT-XL~\cite{ma2024sit} backbone (686.29M) and sinusoidal positional embeddings.
We train unconditional models on SkyTimelapse~\cite{dtvnetsky}, and class-conditional models on UCF-101~\cite{soomro2012ucf101} and Kinetics-600 \cite{kay2017kinetics}.
All models are trained at \(256^2 \times 16\) for 100K iterations with EMA decay \(0.9999\), following~\cite{ma2024sit}.
We report gFVD computed from 5K generated samples.

\textbf{Baselines.}
We compare against tokenizers paired with the same SiT-XL generator to isolate tokenization effects.
ElasticTok requires a per-sample binary search to choose the token budget, which is computationally impractical in our training pipeline; we therefore tokenize non-overlapping 16-frame clips offline.
Since ElasticTok operates on 4-frame chunks, we additionally provide a chunk-index channel.
Due to resource constraints, ElasticTok is evaluated on SkyTimelapse and UCF-101 only.

\textbf{Structure-aware generation under sparsity.}
We study two strategies to mitigate the issue of content--position misalignment.
(i) \emph{Joint generation} concatenates content and position representations channel-wise and applies decoupled timestep schedules; it can reduce misalignment but is sensitive to the choice of the two schedules, requiring tuning.
(ii) \emph{Cascaded generation} predicts a binary selection mask with a lightweight mask prior trained via flow matching, then conditions the main generator on the selected positions through positional embeddings.

During training, the number of active tokens is provided as an additional conditioning signal, allowing the model to learn generation behaviors under varying sparsity levels.
At inference time, adjusting the length of the initial noise tokens enables intuitive modulation of motion complexity and visual detail (Fig.~\ref{fig:generation-ntokens}).

\textbf{Results.}
Our generator operates in a substantially more compact token space: on average we generate 366 tokens per clip, whereas Omni-VAE uses a fixed budget of 5,120 tokens.
Despite this \(\sim\)11\(\times\) reduction in the number of generated tokens, our method consistently outperforms Omni-VAE across all datasets (Table~\ref{tab:generation_fvd}).
Moreover, Table~\ref{tab:gen_ablation_ucf} shows a clear progression in generation quality under the same tokenizer: naive flow matching already surpasses Omni-VAE, joint content--position generation further improves gFVD, and our cascaded mask-prior conditioning yields the largest gain.

\begin{table}[t]
\centering
\footnotesize
\setlength{\tabcolsep}{4pt}
\caption{\textbf{Ablations on VAE components (100K training).}
Higher PSNR/SSIM indicate better reconstruction and lower LPIPS is better.
\texttt{Tokens} reports the average number of active tokens (max = 514).
$^\dagger$ indicates training collapse (only register tokens remain active).}
\label{tab:ablation}
\begin{tabular}{lcccc}
\toprule
\textbf{Configuration} & \textbf{PSNR}$\uparrow$ & \textbf{LPIPS}$\downarrow$ & \textbf{SSIM}$\uparrow$ & \textbf{Tokens}$\downarrow$ \\
\midrule
\textbf{Full model} & 30.85 & 0.07 & 0.93 & 365.57 \\
$-$ latent reg. & 31.26 & 0.07 & 0.94 & 361.70 \\
$-$ asymmetric decoding & 29.61 & 0.11 & 0.92 & 377.80 \\
$-$ Video-LPIPS & 29.28 & 0.11 & 0.91 & 404.00 \\
$-$ register tokens & 29.17 & 0.12 & 0.91 & 410.20 \\
$-$ soft attention mask$^\dagger$ & 19.00 & 0.51 & 0.54 & 2.00 \\
$-$ Gumbel-Softmax$^\dagger$ & 18.91 & 0.52 & 0.53 & 2.00 \\
\bottomrule
\end{tabular}
\end{table}

\begin{table*}[t]
\centering
\footnotesize

\begin{minipage}[t]{0.45\textwidth}
\centering
\setlength{\tabcolsep}{4pt}
\captionof{table}{\textbf{Video generation on \texttt{Sky}, \texttt{UCF}, and \texttt{Kinetics} (gFVD$\downarrow$).} Our method achieves the best performance across all datasets.}
\label{tab:generation_fvd}
\begin{tabular}{lccc}
\toprule
\textbf{Model} & \textbf{Sky} & \textbf{UCF} & \textbf{Kinetics} \\
\midrule
Omni\text{-}VAE & 23.28 & 100.00 & 206.58 \\
Elastic\text{-}KL & 95.53 & 712.56 & --- \\
\midrule
Ours\text{-}Cascaded & 23.19 & \textbf{61.53} & \textbf{160.84} \\
Ours\text{-}Joint & \textbf{21.36} & 73.16 & 193.78 \\
\bottomrule
\end{tabular}
\end{minipage}
\hspace{0.04\textwidth}
\begin{minipage}[t]{0.45\textwidth}
\centering
\setlength{\tabcolsep}{4pt}
\captionof{table}{\textbf{Generation ablation on \texttt{UCF101} (gFVD$\downarrow$).} All use \textbf{Ours} tokenizer. See section \ref{sec:abl} for additional details.}
\label{tab:gen_ablation_ucf}

\begin{tabular}{lc}
\toprule
\textbf{Variant} & \textbf{gFVD}$\downarrow$ \\
\midrule

\multicolumn{2}{l}{\textbf{VAE}} \\
\cmidrule(lr){1-2}
Stage-1 (full) & \textbf{101.23} \\
w/o latent reg. & 161.23 \\

\midrule

\multicolumn{2}{l}{\textbf{Diffusion}} \\
\cmidrule(lr){1-2}
Naïve & 95.69 \\
Joint & 73.16 \\
Cascaded & \textbf{61.53} \\

\bottomrule
\end{tabular}

\end{minipage}
\end{table*}

\begin{figure*}[t]
    \centering
    \includegraphics[width=\textwidth]{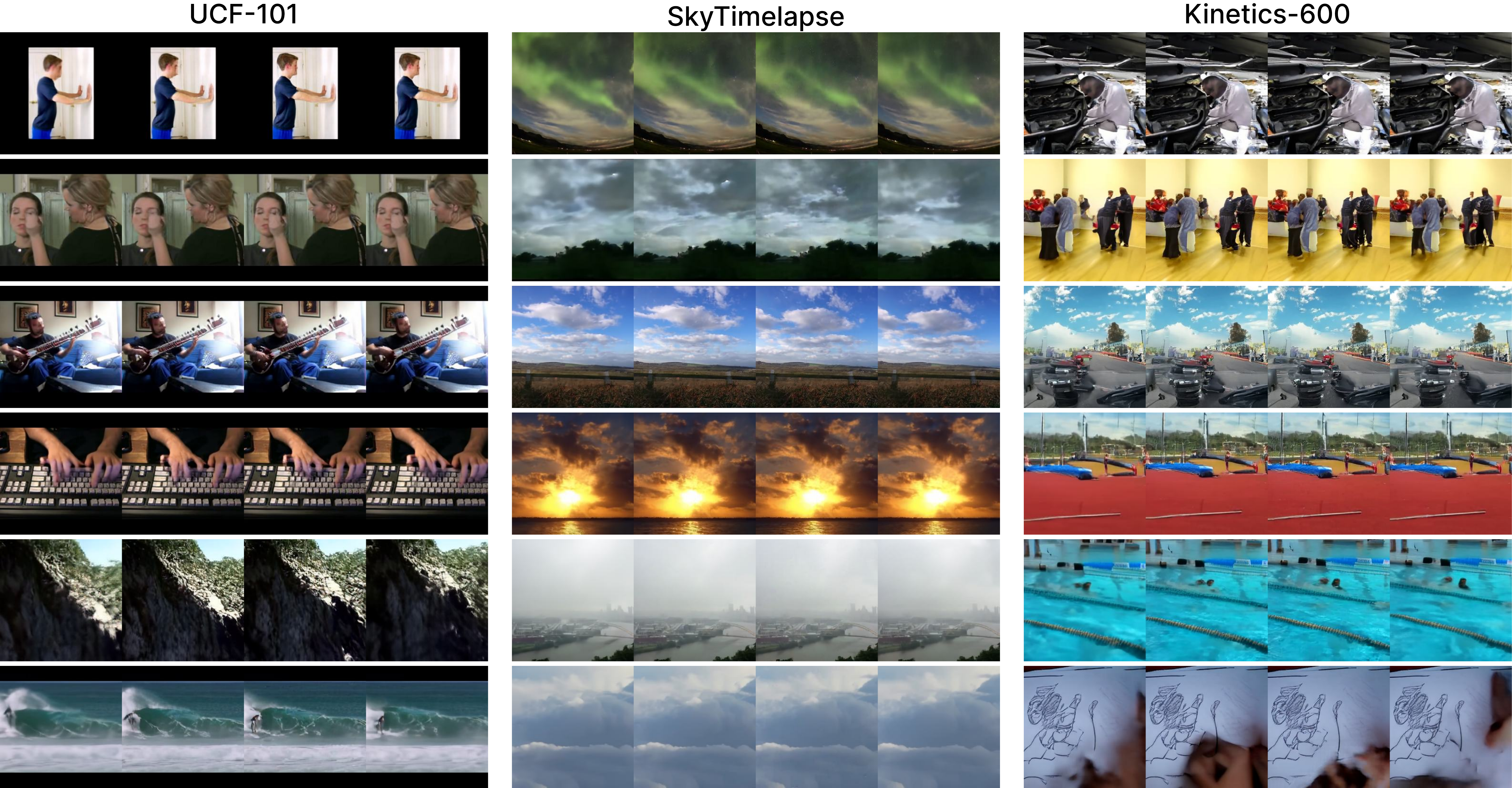}
    \caption{\textbf{Generation quality across datasets.} Qualitative results on UCF-101, Sky-Timelapse, and Kinetics-600.}
    \label{fig:generation}
\end{figure*}

Fig.~\ref{fig:wall_clock} shows that our model achieves superior generation performance in terms of both convergence speed and final quality. In terms of gFVD, Ours-Cascaded reaches \textbf{49.34} at 200k steps compared to 82.31 for OmniTokenizer, a conventional dense Transformer tokenizer, while using only $\sim$28.6\% of the effective tokens on average under OmniTokenizer’s $2\times2$ spatial patch setting. Notably, it achieves 73.81 at 80k steps, surpassing OmniTokenizer’s final result at 200k steps and achieving a $\sim$6.9$\times$ wall-clock speed-up. Compared to ElasticTok, a flexible baseline tokenizer, our model is $\sim$46.7$\times$ faster, reaching 203.51 at 30k steps versus 571.41 at 200k steps. For ElasticTok, results are obtained using pre-computed latents with sequence lengths determined via binary search, and training time is measured under 1-NFE encoding. All models are trained for 200k steps on the UCF-101 dataset and evaluated using EMA weights. In terms of generation throughput, Ours-Cascaded achieves 15.71 videos/sec, achieving 3.1$\times$, 3.2$\times$, and 3.7$\times$ speed-ups over Ours-Joint (5.01), OmniTokenizer (4.91), and ElasticTok (4.24), respectively, using the DOPRI5 ODE solver. All reported training and inference times are measured on 8 H200 GPUs with a global batch size of 256.

Beyond generation quality, the number of tokens naturally functions as a control signal during generation.
Using fewer tokens leads to simpler, low-motion videos, while increasing the token count produces more dynamic and visually detailed sequences.
As illustrated in Fig.~\ref{fig:generation-ntokens}, this emergent controllability allows intuitive modulation of motion intensity and visual richness without retraining or additional conditioning, arising from the statistical tendency of higher token allocations to correlate with more complex scenes, as visualized in Fig.~\ref{fig:entropy}.

Qualitative generation results across UCF-101, SkyTimelapse, and Kinetics-600 are shown in Fig.~\ref{fig:generation}, where our method produces temporally coherent and visually detailed videos across diverse scene types.
Overall, our method achieves superior generation quality in the sparse-token regime, consistently outperforming dense tokenization baselines while generating an order of magnitude fewer tokens. These results indicate that adaptive tokenization is not only computationally efficient, but can also improve generative fidelity.

\subsection{Ablation}
\label{sec:abl}
We conduct controlled ablations under a fixed training budget for fair comparison. Results are summarized in Table~\ref{tab:ablation} and Table~\ref{tab:gen_ablation_ucf}.

\textbf{VAE components (Table~\ref{tab:ablation}).}
We ablate the VAE components using the stage-1 model (single-resolution setting) trained for 100K iterations, where each variant removes or replaces one module from the full configuration. In addition, we evaluate our asymmetric decoding: compared to the $16^2\times8$ decoder-patch model, the full model increases only the decoder reconstruction granularity ($8^2\times4$), improving reconstruction quality substantially (PSNR 29.61$\rightarrow$31.26) while keeping the average token count in a similar range (377.80 vs.\ 361.70).

Replacing Video-LPIPS with an image-based perceptual loss reduces reconstruction quality and increases token usage (377.80$\rightarrow$404.00), suggesting that temporal-aware perceptual supervision improves both coherence and token efficiency. Removing register tokens shows a similar degradation, indicating that they stabilize global structure and improve token utilization.

The soft attention mask is the most critical: removing it collapses training, leaving only the two register tokens active (PSNR 19.00, SSIM 0.54). A comparable collapse occurs without Gumbel--Softmax, confirming that differentiable sampling is essential for learning meaningful token selection.

\textbf{Generation ablations.}
Table~\ref{tab:gen_ablation_ucf} reports gFVD ablations on UCF101 under two controlled settings.

We first isolate the effect of latent regularization using the \emph{stage-1} tokenizer trained for 100K iterations (single-stage setting). While latent regularization slightly compromises reconstruction metrics, it substantially improves downstream generation, as removing it degrades gFVD from 101.23 to 161.23. We then fix the \emph{full} tokenizer setting (all stages) and vary only the diffusion model design. Under the same tokenizer and training protocol, naive content-only flow matching achieves 95.69 gFVD but suffers from content–position misalignment under high sparsity; joint content--position generation improves performance to 73.16, and our cascaded mask-prior conditioning performs best (61.53 gFVD), supporting the benefit of decoupling position selection from content generation.
\section{Conclusion}
\label{sec:conclusion}
This work introduces \textbf{KATok}, an adaptive VAE that selectively keeps or drops 
tokens according to content complexity for compact and expressive video representation. 
Through differentiable attention gating and sparsity loss, the adaptive token selector 
achieves an effective balance between fidelity and efficiency. 
To address the content--position misalignment induced by sparse latent tokens, 
we propose two remedies: joint content--position generation with timestep decoupling 
and a cascaded mask-prior conditioning scheme. 
Together, these results demonstrate that adaptive tokenization, when coupled 
with suitable diffusion design, provides a principled framework for 
efficient and high-quality video generation.

\section*{Acknowledgements}
We thank Jisu Choi for valuable research discussions and insightful feedback, and Hansaem Kim for support and encouragement throughout this project.

%
%
\bibliographystyle{splncs04}
\bibliography{main}
\end{document}


\title{\texorpdfstring{Supplementary Material for ``Keep-or-Drop? Adaptive Tokenizer for Compact Video Representation''}
{Supplementary Material}}

\author{Yeonkyeong Lee \and
Hyunsung Go \and
Jongmin Kim \and
Sewoong Lim \and
Donghoon Lee\thanks{Corresponding author.}}
\institute{Kakao Corp., Republic of Korea\\
\email{\{mag.ic, cog.map, lukas.ai, amita.lim, dev.e\}@kakaocorp.com}}
\maketitle


\section{Implementation Details}

\begingroup
\scriptsize
\setlength{\tabcolsep}{2pt}
\renewcommand{\arraystretch}{1.08}

\begin{longtable}{
  >{\raggedright\arraybackslash}p{0.34\textwidth}
  >{\raggedright\arraybackslash}p{0.19\textwidth}
  >{\raggedright\arraybackslash}p{0.19\textwidth}
  >{\raggedright\arraybackslash}p{0.19\textwidth}
}
\caption{\textbf{Training configurations for the three-stage training pipeline of KATok.}
Stage 1 performs VAE pretraining with pixel- and perceptual-level supervision.
Stage 2 extends training to multi-resolution video data for enhanced spatial--temporal generalization.
Stage 3 introduces adversarial fine-tuning with non-saturating GAN loss to improve perceptual sharpness and realism.}
\label{supp:configuration} \\
\toprule
\textbf{Component} & \textbf{Stage 1} & \textbf{Stage 2} & \textbf{Stage 3} \\
\midrule
\endfirsthead

\toprule
\textbf{Component} & \textbf{Stage 1} & \textbf{Stage 2} & \textbf{Stage 3} \\
\midrule
\endhead

\midrule
\multicolumn{4}{r}{\textit{(Continued on next page)}} \\
\endfoot

\bottomrule
\endlastfoot

\multicolumn{4}{l}{\textbf{Model Architecture}} \\
\multicolumn{4}{l}{\textit{(All resolution tuples are [H, W, T].)}} \\
Encoder patch size & [16,16,8] & [16,16,8] & [16,16,8] \\
Decoder patch size & [8, 8, 4] & [8, 8, 4] & [8, 8, 4] \\
Hidden dimension & 768 & 768 & 768 \\
Number of heads & 16 & 16 & 16 \\
MLP ratio & 4.0 & 4.0 & 4.0 \\
Dropout ($p$) & 0.0 & 0.0 & 0.0 \\
Latent channels & 64 & 64 & 64 \\
Encoder \# layers & 12 & 12 & 12 \\
Decoder \# layers & 18 & 18 & 18 \\
Register tokens & 2 & 2 & 2 \\
Weight initialization & Truncated normal & -- & -- \\

\midrule
\multicolumn{4}{l}{\textbf{Gumbel--Softmax Settings}} \\
Initial $\tau$ & 2.0 & 0.1 & 0.1 \\
Minimum $\tau$ & 0.1 & 0.1 & 0.1 \\
Anneal steps & 20K & -- & -- \\
Min keep probability & 0.01 & 0.01 & 0.01 \\

\midrule
\multicolumn{4}{l}{\textbf{Loss Configuration}} \\
Loss functions used &
\makecell[l]{L1\\Video-LPIPS\\Sparse\\KLD\\vJEPA-2 align} &
\makecell[l]{L1\\Video-LPIPS\\Sparse\\KLD\\vJEPA-2 align} &
\makecell[l]{L1\\Video-LPIPS\\Sparse\\KLD\\vJEPA-2 align\\Adversarial} \\
\noalign{\vskip 2.0ex}

L1 loss weight & 1.0 & 1.0 & 1.0 \\
Video-LPIPS weight & 0.1 & 0.1 & 0.1 \\
Sparse loss weight & 0.01 & 0.01 & 0.01 \\
Sparse loss start step & 5K & -- & -- \\
Sparse loss anneal step & 20K & -- & -- \\
KLD loss weight & $1\times 10^{-7}$ & $1\times 10^{-7}$ & $1\times 10^{-7}$ \\
vJEPA-2 align & 0.5 & 0.5 & 0.5 \\
vJEPA-2 align \# layers & 2 & 2 & 2 \\
Noise aug sigma & 0.2 & 0.2 & 0.2 \\

\midrule
\multicolumn{4}{l}{\textbf{Adversarial Loss (Stage 3 only)}} \\
Generator loss weight & -- & -- & 0.1 \\
Discriminator loss weight & -- & -- & 1.0 \\
\noalign{\vskip 2.0ex}

Loss type & -- & -- & \makecell[l]{Non-saturating\\GAN loss} \\
\noalign{\vskip 2.0ex}

Gen/Disc anneal steps & -- & -- & 2K \\
Approx.\ R1/R2 weight $\gamma$ & -- & -- & 0.25 \\
Approx.\ R1/R2 $\epsilon$ & -- & -- & $1\times 10^{-3}$ \\

\midrule
\multicolumn{4}{l}{\textbf{Training Setup}} \\
Effective batch size & 256 & 256 & 256 \\
Input shapes & [256,256,16] &
\makecell[l]{Multi-res\\(see Sec.~\ref{para:multi_stage_train})} &
\makecell[l]{Same as\\Stage 2} \\
\noalign{\vskip 2.0ex}

Total steps & 210K & 30K & 50K \\
\noalign{\vskip 2.0ex}

Optimizer & AdamW & AdamW & \makecell[l]{AdamW (Gen)\\Adam (Disc)} \\
\noalign{\vskip 2.0ex}

Warm-up steps & 2K & 2K & \makecell[l]{500 (Gen)\\500 (Disc)} \\
\noalign{\vskip 2.0ex}

Learning rate & $5\times 10^{-5}$ & $5\times 10^{-5}$ &
\makecell[l]{$1\times 10^{-6}$ (Gen)\\$5\times 10^{-5}$ (Disc)} \\

\noalign{\vskip 2.0ex}
Betas & [0.9,0.999] & [0.9,0.999] &
\makecell[l]{[0.5,0.999] (Gen)\\{}[0.0,0.9] (Disc)} \\
\noalign{\vskip 2.0ex}

Weight decay & 0.0001 & 0.0001 & \makecell[l]{0.0001 (Gen)\\0 (Disc)} \\
\noalign{\vskip 2.0ex}

Scheduler & Cosine annealing & Cosine annealing & Cosine annealing \\
EMA decay & 0.9999 & 0.9999 & 0.9999 \\
Precision & 32-true & 32-true & 32-true \\
Grad clip (norm) & 1.0 & 1.0 & 1.0 \\

\end{longtable}
\endgroup

\subsection{Tokenizer Model Architecture.}
The proposed KATok model is a transformer-based adaptive VAE composed of a linear patchifier/unpatchifier, a single-stream encoder, a double-stream decoder, and the adaptive token selector.
The patchifier and unpatchifier are implemented as linear projection layers.
The encoder patchifier uses scale factors of $16^2 \times 8$ (spatial $\times$ temporal), while the decoder unpatchifier operates at a finer scale of $8^2 \times 4$ to enable asymmetric coarse-to-fine decoding (detailed below).
The encoder follows the single-stream block design of FLUX~\cite{flux2024} but removes the modulation layer for simplicity, and applies 3D rotary positional embeddings (3D RoPE~\cite{su2024rope}) to encode spatio-temporal information.
The model specification generally follows the ViT-B configuration (hidden dim 768, MLP ratio 4.0), with 16 attention heads and a total of 344M parameters. Full configurations are shown in Table~\ref{supp:configuration}.

The decoder adopts the double-stream architecture of FLUX without modulation.
One stream accepts learnable query tokens $\{q_j\}_{j=1}^{M}$ that are repeated to match the shape of the patchified video input (where $M$ equals the fine-grid patch count), and applies 3D RoPE to reconstruct spatial layouts, while the other stream processes the latent tokens without any positional embeddings, by assigning all-zero position indices.
A soft attention mask is applied to the latent stream of the decoder’s double-stream block to maintain consistent token influence across layers, and layer normalization is applied to the latent tokens before entering the decoder stack.
To avoid materializing a full $N{\times}N$ attention mask matrix to ensure compatibility with FlashAttention~\cite{dao2022flashattention}, we employ a \emph{KV bias trick}: the query and key embeddings are augmented as $q_i \leftarrow [q_i, 1], k_j \leftarrow [k_j, \sqrt{d} b_j]$, where $b_j = \log(\widetilde{m}_j + \varepsilon)$, as defined in the main paper, and $d$ denotes the dimensionality, yielding equivalent masking behavior with substantially lower memory overhead.

The adaptive token selector receives the encoder output $e_i$ and maps it into the parameters of a diagonal Gaussian $(\mu_i, \sigma_i) = f_\theta(e_i)$, as well as Gumbel keep/drop logits $\alpha_i = g_\theta(e_i) \in \mathbb{R}^2$, following the notation in the main paper.
The mask values corresponding to register tokens are always set to $1$, ensuring they remain active throughout training.
We clamp soft mask values~$<0.01$ to~$0.0$, and use Gumbel-Softmax with $\tau$ annealed from 2.0 to 0.1 over 10K steps during stage 1, always without discretization (i.e., with hard=False).
The overall loss function combines L1, video-LPIPS, sparsity, KLD, latent regularizations, and adversarial losses with stage-specific weights shown in Table~\ref{supp:configuration}. The KLD loss is defined as a weighted sum of token-wise KLD losses with respect to the unit Gaussian prior, with the corresponding soft masks $\widetilde{m}_i$ as weights.
Both sparsity and KLD losses are linearly annealed, introduced after 5K iterations and reaching full weight by 20K in stage 1.

\vspace{2mm}
\paragraph{Asymmetric Coarse-to-Fine Decoding.}
The encoder and decoder operate at different patch granularities: the encoder uses coarse patches of size $16^2 \times 8$, while the decoder reconstructs at a finer grid of $8^2 \times 4$.
The coarse encoder patch size ($16^2 \times 8$) reduces the total number of tokens, which is desirable for compact representation but inevitably discards fine-grained spatial--temporal details.
To compensate for this information loss, the decoder operates at a finer patch granularity ($8^2 \times 4$), allowing it to recover high-frequency details from the compact latent set.
In practice, the encoder patchifies a $256^2{\times}16$ video into $16^2{\times}2 = 512$ tokens, while the decoder uses $32^2{\times}4 = 4{,}096$ learnable query tokens.
Only the number of query tokens changes; the latent token set remains compact, preserving computational efficiency during both training and generation.
Removing this asymmetric design (i.e., using identical patch sizes for both encoder and decoder) degrades reconstruction quality (PSNR 31.26 $\to$ 29.61), confirming that finer decoder resolution is essential for high-fidelity reconstruction under aggressive token sparsification.

\vspace{2mm}
\paragraph{Latent Regularization.}
We apply two complementary regularization techniques to the latent tokens to stabilize training and improve generative quality.

\noindent\textit{(i) Latent noise augmentation.}
During training, we inject a small amount of Gaussian noise into the sampled latent tokens before they enter the decoder \cite{hacohen2024ltx}.
Specifically, for each sample in a batch we draw a noise level $\eta \sim \mathcal{U}(0, \sigma)$ with $\sigma = 0.2$ and perturb the latent as
$\tilde{z}_i = (1-\eta)\,z_i + \eta\,\epsilon$, where $\epsilon \sim \mathcal{N}(0,\mathbf{I})$.
This stochastic corruption acts as a form of input-space regularization for the decoder, preventing it from memorizing exact encoder outputs and encouraging robustness to perturbations in the latent space.

\noindent\textit{(ii) Representation alignment with vJEPA-2.}
We align the VAE's latent representations with those of a frozen vJEPA-2 ViT-H encoder~\cite{assran2025v} to encourage semantically meaningful latent codes.
The alignment is realized through a lightweight auxiliary decoder $h_\psi$ and a 3D pixel-shuffle projector.

Concretely, the auxiliary decoder $h_\psi$ is a 2-layer transformer that mirrors the main decoder's double-stream block design (with soft attention masking and latent normalization), but operates at a coarser resolution of $16^2 \times 2$ (i.e., $512$ query tokens).
It takes the sparse latent tokens and produces coarse-grid features of shape $(B, 2 \times 16 \times 16, d)$, where $d = 768$ is the hidden dimension.

These features are then upsampled to match the vJEPA-2 native token grid of $16^2 \times 8$ ($2{,}048$ tokens) via a 3D pixel-shuffle projector.
The projector first maps each token through a LayerNorm--Linear--GELU bottleneck (bottleneck dimension $= 768$), then rearranges the temporal dimension with a scale factor of $4{\times}$ (upsampling from 2 to 8 along the temporal axis), and finally applies a linear projection to the vJEPA-2 feature dimension ($1{,}280$).

On the target side, the input video is resized to $256 \times 256$ and processed by the frozen vJEPA-2 encoder to obtain patch-level features of shape $(B, 2048, 1280)$.
The alignment loss is then defined as the mean cosine dissimilarity between the projected VAE features and the vJEPA-2 features:
\[
\mathcal{L}_{\text{align}} = 1 - \frac{1}{N_q} \sum_{j=1}^{N_q} \frac{\hat{y}_j \cdot y_j}{\|\hat{y}_j\| \, \|y_j\|},
\]
where $\hat{y}_j$ and $y_j$ denote the projected VAE feature and frozen vJEPA-2 feature at position $j$, respectively, and $N_q = 2{,}048$.
This loss is weighted by $\lambda_{\text{align}} = 0.5$ in all stages.
Note that the mask values $\widetilde{m}_i$ are \emph{detached} before entering the auxiliary decoder, so that $\mathcal{L}_{\text{align}}$ does not push mask values toward 1 and thus does not conflict with the sparsity loss.

\vspace{2mm}
\paragraph{Multi-Stage Training.}
\label{para:multi_stage_train}
While the model supports end-to-end optimization, we adopt a three-stage training schedule for computational efficiency.
In \textbf{Stage~1}, the model is trained for 210K iterations at a single resolution of $256^2\times16$ frames to learn stable reconstruction and token sparsity.
In \textbf{Stage~2}, the model undergoes 30K iterations of multi-resolution training to improve generalization across different spatio-temporal scales, using input configurations:
\[
\begin{aligned}
&\text{[256, 256, 16], [256, 256, 24], [256, 256, 32], [256, 256, 64],}\\
&\text{[512, 512, 8], [512, 512, 16], [368, 640, 16], [640, 368, 16],}\\
&\text{[240, 416, 32], [416, 240, 32].}
\end{aligned}
\]
These resolutions are uniformly sampled per iteration to balance temporal and spatial diversity. 
Finally, \textbf{Stage~3} performs 50K iterations of adversarial fine-tuning with a reconstruction GAN to enhance perceptual fidelity and sharpness.

All stages were trained on a multi-node GPU cluster, with each node equipped with eight NVIDIA H200 GPUs (140 GB memory each), using an effective batch size of 256.

\vspace{2mm}
\paragraph{Stage 3: GAN Training.}
We adopt a reconstruction-based GAN framework inspired by LTX-Video~\cite{hacohen2024ltx}, extending it by initializing the discriminator with the weights of the pretrained encoder from Stage~2.
At initialization, the patchifier is replicated to accept two video inputs (real and fake) simultaneously. To mitigate activation shifts introduced by the added patchifier, we apply an element-wise affine transformation to the replicated patchifier, with the parameters zero-initialized to stabilize the early phase of training.
The discriminator extracts register-token embeddings for discrimination, using intermediate features from layers~3,~9, and~11 (0-based indexing). These embeddings are concatenated and projected linearly to produce the final logit.
As we work within the framework of reconstruction GANs, we define the order \([ \text{real}, \text{fake} ]\) as the ``real'' input and the reverse order \([ \text{fake}, \text{real} ]\) as the ``fake'' input to establish the logit semantics.
The generator (encoder--decoder) is trained jointly with the discriminator under the non-saturating GAN losses, with both losses linearly annealed over 2K iterations as summarized in Table~\ref{supp:configuration}.
We also apply the approximated R1/R2 regularization from \cite{pmlr-v267-lin25m, huang2024r3gan} with a regularization weight~$\gamma=0.25$ and~$\epsilon = 1\times10^{-3}$, corresponding to~$\frac{\lambda}{2\sigma}$ and~$\sigma$ in~\cite{pmlr-v267-lin25m}, respectively, for both R1 and~R2 to stabilize adversarial updates.

\vspace{2mm}
\noindent Overall, this architecture and training strategy allow KATok to efficiently learn adaptive video representations while preserving spatial–temporal coherence under aggressive token sparsification.

\subsection{Diffusion Model Architecture.}
\label{sec:diffusion_arch}

\paragraph{Content Diffusion Model.}
Because the adaptive tokenizer produces a \emph{variable-length} subset of tokens whose spatial positions are non-uniform, the diffusion model must handle two challenges absent in dense-token settings: (i)~the number of tokens varies across samples, and (ii)~the standard grid-aligned positional encoding no longer applies since dropped tokens create gaps in the spatial layout.
We address these with two generation variants—\emph{joint} and \emph{cascaded}—that differ in how token positions are represented (detailed below).

The generation model follows the SiT-XL~\cite{ma2024sit} architecture: depth~28, hidden dimension~1152, 16~attention heads, MLP ratio~4.0, and QK-normalization enabled, totaling 686.29M parameters.
Each block uses adaptive layer normalization zero (adaLN-Zero) for timestep and class conditioning.
The positional embedding differs between the two variants.
In the \emph{joint} variant, 1D sinusoidal positional embeddings encode the packing order of tokens within each sample (not spatial positions), while explicit spatial information is provided by concatenating raw 3D coordinates to the content channels.
In the \emph{cascaded} variant, no 1D sequential embedding is used; instead, ground-truth 3D token coordinates $(t,h,w)$ (snapped to the nearest grid-cell centers) are mapped to the hidden dimension via a two-layer MLP with SiLU activation, providing explicit spatial conditioning.
At inference time, the same MLP converts coordinates predicted by the mask prior into positional embeddings.
Both variants share the same three conditioning signals combined additively via AdaLN-Zero: (i)~a sinusoidal timestep embedding, (ii)~a class label embedding with 10\% classifier-free guidance dropout, and (iii)~a token-length embedding that encodes the target number of active tokens via sinusoidal encoding, also with 10\% dropout.
The token-length conditioning enables explicit control over the sparsity level at generation time.
For all comparative experiments, the same diffusion backbone is used across models, differing only in the VAE encoder that provides the latent representations. All models are trained at a resolution of $256^2 \times 16$.

\vspace{2mm}
\paragraph{Training Configuration.}
Table~\ref{tab:diffusion_config} summarizes the training configurations for the two generation variants described in the main paper: \emph{joint content--position generation} and \emph{cascaded generation}.
Both variants share the same SiT-XL backbone for content generation, trained with AdamW at a constant learning rate of $1 \times 10^{-4}$ and EMA decay of 0.9999 for 100K iterations with a global batch size of 256.
Since the number of tokens varies across samples, we pack all tokens from a batch into a single concatenated sequence and employ the \emph{varlen} variant of FlashAttention~\cite{dao2022flashattention}, which processes the packed sequence with per-sample boundaries, avoiding both explicit attention masking and padding overhead.
Latent tokens are normalized using precomputed channel-wise mean and standard deviation before being fed to the diffusion model, and denormalized before VAE decoding.

\begin{table}[t]
\centering
\footnotesize
\caption{\textbf{Diffusion training configurations.}
Both generation variants share the same SiT-XL content backbone (top).
The \emph{joint} variant concatenates 3D position coordinates to the content channels;
the \emph{cascaded} variant instead adds a lightweight mask prior module that is trained jointly with the content model.}
\label{tab:diffusion_config}
\begin{tabular}{ll}
\toprule
\multicolumn{2}{l}{\textbf{Shared content model (SiT-XL)} --- used by both variants} \\
\midrule
Architecture / Parameters & SiT-XL / 686.29M \\
Hidden dim / Depth / Heads & 1152 / 28 / 16 \\
Input channels & 64 (latent) \\
QK-normalization & \checkmark \\
Positional embedding & Variant-dependent (see text) \\
Optimizer & AdamW ($\beta_1{=}0.9$, $\beta_2{=}0.999$), wd $= 0$ \\
Learning rate & $1 \times 10^{-4}$ (constant) \\
EMA decay & 0.9999 \\
Batch size / Iterations & 256 / 100K \\
Token-length conditioning & Sinusoidal embedding (10\% CFG dropout) \\
\midrule
\multicolumn{2}{l}{\textbf{Joint variant} --- adds position channels to the content model} \\
\midrule
Additional input channels & +3 (raw $(t,h,w)$ coordinates), total 67 \\
Noise schedule & Decoupled ($t_{\text{content}}$, $t_{\text{pos}}$ sampled independently) \\
Position loss weight & $\lambda_{\text{pos}} = 1.0$ \\
Position noise dropout & 10\% (for position-branch CFG) \\
\midrule
\multicolumn{2}{l}{\textbf{Cascaded variant} --- adds a separate mask prior module} \\
\midrule
Architecture / Parameters & SiT (custom) / 8.3M \\
Hidden dim / Depth / Heads & 192 / 12 / 3 \\
Input & 1-channel mask over $2{\times}16{\times}16$ grid \\
QK-normalization & \checkmark \\
Param group & Separate group (wd $= 0.01$), shared AdamW optimizer \\
Learning rate & $1 \times 10^{-4}$ (constant) \\
EMA decay & 0.9999 \\
Prior loss weight & $\lambda_{\text{prior}} = 0.1$ \\
Training & Jointly trained with content model \\
\bottomrule
\end{tabular}
\end{table}

In contrast to our model, OmniTokenizer and ElasticTok produce substantially more tokens due to their finer patch sizes.
To maintain computational feasibility for diffusion training, we apply an additional patchification, yielding a 4$\times$ reduction in token count (OmniTokenizer: $2{\times}2$ spatial patches; ElasticTok: 1D patch size of 4), thereby reducing the effective token count for the DiTs.
ElasticTok also performs variable-length tokenization,
but its token count is determined on a per-chunk basis (4-frame units).
To enable decoding into videos,
we append an additional channel to the latent tokens that marks the last token of each chunk with 1 and all others with 0, then concatenate all chunks into a single sequence for training.

\vspace{2mm}
\paragraph{Joint Content--Position Generation.}
As described in the main paper, the joint approach augments the SiT-XL input by concatenating normalized 3D spatial coordinates $(t, h, w) \in \mathbb{R}^3$ to each latent token, increasing the per-token channel dimension from 64 to 67.
Content and position are diffused through fully decoupled noise schedules: at each training step, independent timesteps $t_{\text{content}}$ and $t_{\text{pos}}$ are sampled from $\mathcal{U}(0,1)$.
During training, the concatenated representation is processed jointly within the diffusion network; the output is then separated into position and content branches, optimized with independent flow-matching losses $\mathcal{L}_{\text{content}}$ and $\mathcal{L}_{\text{pos}}$, both weighted equally (i.e., $\lambda_{\text{pos}} = 1$).
To enable classifier-free guidance on the position branch, we apply \emph{position noise-dropout}: with 10\% probability per sample, the position channels are replaced with pure Gaussian noise, allowing the model to generate positions unconditionally at inference time.

\vspace{2mm}
\paragraph{Cascaded Generation Model.}
The cascaded model first generates an occupancy mask over a fixed 3D spatio-temporal grid via a lightweight \emph{mask prior}, from which token positions are extracted for the content diffusion model.
The mask prior is a flow-matching generative model implemented as a SiT transformer with the configuration shown in Table~\ref{tab:diffusion_config}.
It is trained jointly with the content diffusion model within a single optimizer, using a separate parameter group with weight decay $0.01$; its loss is weighted by $\lambda_{\text{prior}} = 0.1$.
During training, the content model is conditioned on ground-truth token positions obtained from the encoder; at inference time, these are replaced by the positions predicted by the mask prior.

\noindent\textit{Grid and mask representation.}
We define a fixed 3D grid of size $G = G_t \times G_h \times G_w = 2 \times 16 \times 16 = 512$ cells, whose centers are uniformly spaced in $[-1, 1]^3$.
Given the ground-truth token positions from the encoder, we construct a binary occupancy mask $\mathbf{m} \in \{-1, +1\}^{G}$, where $m_g = +1$ if any token maps to grid cell $g$, and $m_g = -1$ otherwise.
The mask is flattened into a 1D sequence of length $G$ and serves as the generation target.

\noindent\textit{Flow matching objective.}
The prior is trained with a flow-matching loss.
Given a ground-truth mask $\mathbf{m}_0$ and noise $\mathbf{m}_1 \sim \mathcal{N}(0, \mathbf{I})$, we form the interpolant $\mathbf{m}_t = (1-t)\,\mathbf{m}_0 + t\,\mathbf{m}_1$ for $t \sim \mathcal{U}(0,1)$, and train the model to predict the velocity $\mathbf{v}_\theta(\mathbf{m}_t, t, \mathbf{c}) \approx \mathbf{m}_1 - \mathbf{m}_0$.
The loss is $\mathcal{L}_{\text{prior}} = \|\mathbf{v}_\theta - (\mathbf{m}_1 - \mathbf{m}_0)\|^2$.

\noindent\textit{Positional encoding.}
Fixed 3D sinusoidal positional embeddings are computed from the grid coordinates and added to the input sequence, enabling the model to reason about the spatial structure of the mask.

\vspace{2mm}
\paragraph{Inference.}
All models use the Dormand--Prince adaptive ODE solver (\texttt{dopri5}) with absolute tolerance $10^{-6}$ and relative tolerance $10^{-3}$, using 50 interpolation steps.
Classifier-free guidance is applied with a scale of 4.0 for class-label conditioning.

\noindent\textit{Joint variant.}
Noise of shape $(N, 67)$ is sampled from $\mathcal{N}(0, \mathbf{I})$ and integrated from $t=1$ to $t=0$ via a single ODE call.
Although content and position channels share the same state vector, they evolve on \emph{independent time schedules} implemented through logit-normal timestep shifting:
$t' = \mathrm{sigmoid}\bigl(\mathrm{logit}(t) \cdot \sigma_s + \mu_s\bigr)$,
where $(\mu_s, \sigma_s) = (0.0, 1.0)$ for content (identity mapping) and $(-2.0, 0.3)$ for position.
Inside the ODE drift, the model receives the shifted timesteps $t_{\text{content}}$ and $t_{\text{pos}}$ separately, and its velocity output is rescaled by the chain-rule factor $\mathrm{d}t'/\mathrm{d}t$ per channel group so that each group effectively follows its own schedule.
The position shift biases position channels toward earlier completion, reflecting that spatial layout is resolved before fine content details.
The output is split into content (first 64 channels) and position (last 3 channels); content tokens are denormalized and decoded by the VAE.

\noindent\textit{Cascaded variant.}
Inference proceeds in two stages.
First, the cascaded model's mask prior samples Gaussian noise $\mathbf{m}_1 \sim \mathcal{N}(0, \mathbf{I})$ of shape $(G,)$ and solves the ODE $\mathrm{d}\mathbf{m}/\mathrm{d}t = \mathbf{v}_\theta(\mathbf{m}_t, t, \mathbf{c})$ from $t=1$ to $t=0$ to obtain a mask.
We select the top-$k$ grid cells by mask value, where $k$ equals the desired token count, and use their grid-center coordinates as positional conditioning for the content diffusion model.
The content model then generates latent tokens conditioned on these positions, which are decoded by the VAE to produce the final video.

\begin{figure}[t]
    \centering
    \includegraphics[width=\textwidth]{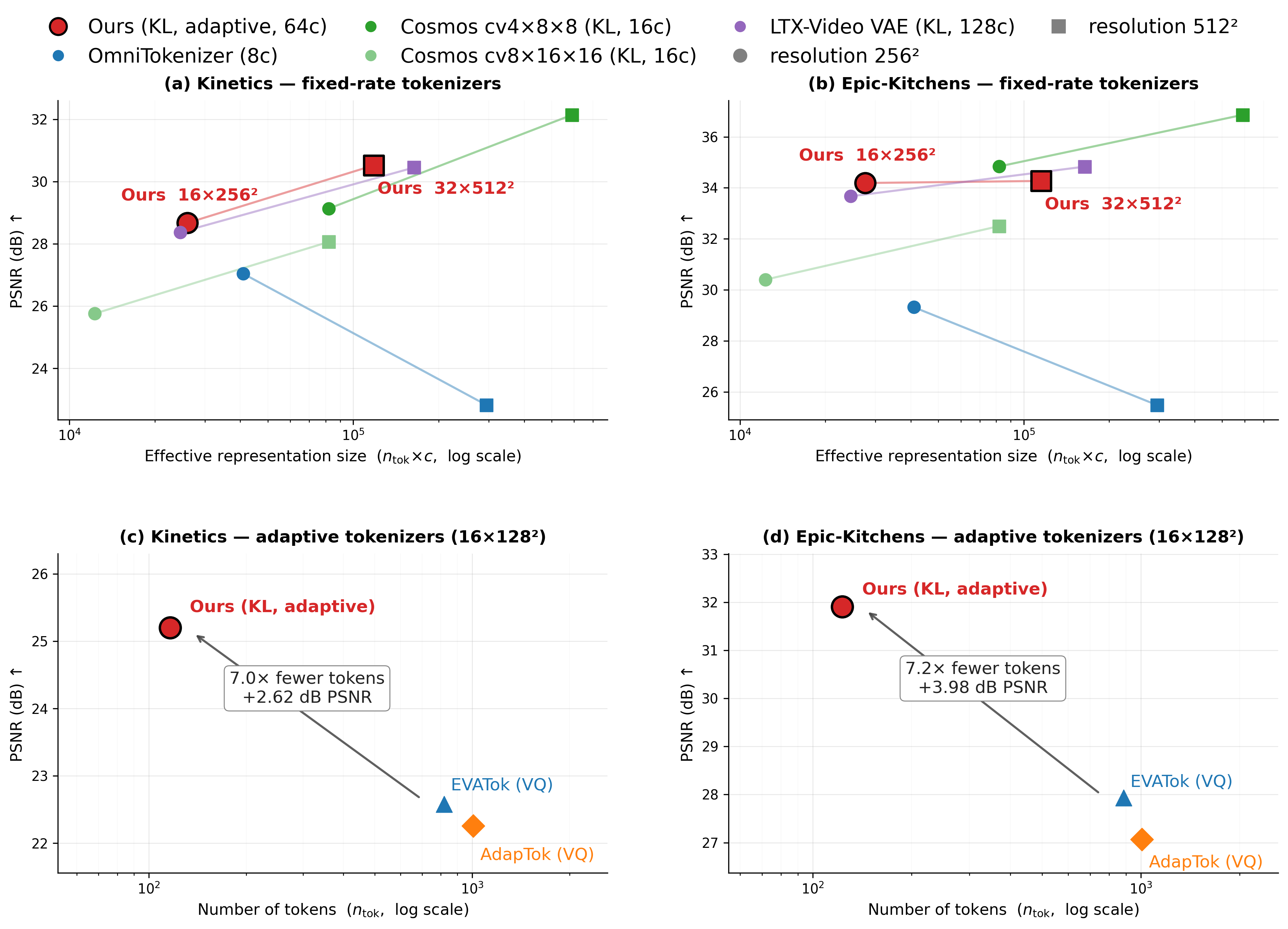}
    \caption{\textbf{Cross-paradigm reconstruction comparison.}
    (a,b) Continuous-token VAEs on Kinetics-600 and Epic-Kitchens at $256^2/512^2$;
    transformer-based (KATok, OmniTokenizer) and CNN-based (LTX-Video, Cosmos) baselines
    all use \emph{fixed} compression, while KATok adapts per-sample. PSNR is plotted against
    total latent size ($\#\text{tokens}\times\text{channels}$) to account for latent-channel
    differences. (c,d) Adaptive VQ tokenizers (EVATok, AdapTok) at $128^2$. KATok remains
    Pareto-competitive across paradigms, and at $128^2$ uses 7$\times$ fewer tokens while
    improving PSNR by 2.6--4.0\,dB.}
    \label{fig:suppl_cross_recon}
\end{figure}

\section{Additional Quantitative Results}

\paragraph{Cross-Paradigm Reconstruction Comparison.}
While the main paper compares KATok only against baselines that share its modeling
paradigm (transformer-based continuous-token VAEs), here we extend the comparison
across paradigms, shown in Fig.~\ref{fig:suppl_cross_recon}.
We evaluate on Kinetics-600~\cite{kay2017kinetics} and
Epic-Kitchens~\cite{damen2022epickitchens}; the latter is an egocentric, head-mounted
stress test whose constant ego-motion leaves few static patches, an unfavorable setting
for adaptive allocation.
Among continuous-token VAEs (a,b), KATok and OmniTokenizer are transformer-based,
whereas LTX-Video~\cite{hacohen2024ltx} and Cosmos~\cite{agarwal2025cosmos} are
CNN-based; all three baselines use fixed compression.
PSNR is plotted against total latent size ($\#\text{tokens}\times\text{channels}$),
consistent with the channel-aware compression ratio used in the main paper.
KATok remains Pareto-leading on Kinetics-600 and Pareto-competitive on the harder
$512^2$ Epic-Kitchens.
We further compare against adaptive \emph{VQ} tokenizers, EVATok~\cite{xiong2026evatok}
and AdapTok~\cite{li2025adaptok}, as cross-paradigm references at $128^2$ (c,d).
Comparing token count against reconstruction quality, KATok reconstructs with
substantially higher fidelity while using $7\times$ fewer tokens, with PSNR gains of
$2.6$--$4.0$\,dB across Kinetics-600 and Epic-Kitchens.

\paragraph{Cross-Paradigm Generation Comparison.}
Table~\ref{tab:suppl_ucf_gen} reports a system-level comparison on UCF-101
class-conditional generation. We group methods by spatial resolution, as prior works
operate at different resolutions and tokenizer paradigms. The bottom group is a controlled
comparison in which all tokenizers are paired with an identical SiT-XL backbone and
training protocol, isolating the effect of the tokenizer alone: KATok attains $61.53$ gFVD
with only $366$ tokens, compared to OmniTokenizer-VAE ($5{,}120$ tokens, $100.00$) and
ElasticTok-KL ($3{,}845$ tokens, $712.56$). The remaining entries (EVATok, AdapTok,
SweetTok, MAGVIT-v2) use VQ tokenizers with AR/MLM generators, and we report numbers from
their respective papers for completeness.

\begin{table}[t]
  \centering
  \footnotesize
  \caption{\textbf{System-level comparison on UCF-101 class-conditional generation.}
  Methods grouped by spatial resolution. Bottom group: controlled comparison with
  identical SiT-XL backbone and training protocol, isolating the effect of tokenization
  design. ``Tok.\ Type'' describes the tokenizer (VQ/KL) and adaptive nature; ``Gen.''\
  denotes generator type.}
  \label{tab:suppl_ucf_gen}
  \setlength{\tabcolsep}{4pt}
  \resizebox{\linewidth}{!}{%
  \begin{tabular}{lllrrr}
  \toprule
  \textbf{Method} & \textbf{Tok.\ Type} & \textbf{Gen.} & \textbf{Gen.\ Params} & \textbf{\#Tokens} & \textbf{gFVD} $\downarrow$ \\
  \midrule
  \multicolumn{6}{l}{\textit{128$^2$ resolution (as reported in prior work)}} \\
  \midrule
  MAGVIT-v2~\cite{yu2024magvitv2} & VQ, fixed & Masked Language Model & 307M & 1{,}280 & 58 \\
  EVATok~\cite{xiong2026evatok} & VQ, adaptive & AR & 633M & 758 & 48 \\
  AdapTok~\cite{li2025adaptok} & VQ, adaptive & AR & 633M & 1{,}024 & 67 \\
  \midrule
  \multicolumn{6}{l}{\textit{256$^2$ resolution (as reported in prior work)}} \\
  \midrule
  OmniTokenizer~\cite{wang2024omnitokenizer} & VQ, fixed & AR & 650M & 5{,}120 & 191 \\
  SweetTok~\cite{tan2025sweettok} & VQ, fixed & AR & 650M & 1{,}280 & 84 \\
  SweetTok (large)~\cite{tan2025sweettok} & VQ, fixed & AR & 1.9B & 1{,}280 & 65 \\
  \midrule
  \multicolumn{6}{l}{\textit{Controlled comparison: same SiT-XL backbone, identical training protocol (256$^2$)}} \\
  \midrule
  OmniTokenizer-VAE~\cite{wang2024omnitokenizer} & VAE, fixed & SiT-XL (Diffusion) & 686M & 5{,}120 & 100.00 \\
  ElasticTok-KL~\cite{yan2024elastictok} & KL adaptive & SiT-XL (Diffusion) & 686M & 3{,}845 & 712.56 \\
  \rowcolor{gray!12}
  \textbf{KATok (Ours)} & \textbf{KL adaptive} & \textbf{SiT-XL (Diffusion)} & \textbf{686M} & \textbf{366} & \textbf{61.53} \\
  \bottomrule
  \end{tabular}%
  }
\end{table}

\section{Additional Qualitative Results}

\begin{figure}[!htbp]
    \centering
    \includegraphics[width=\textwidth]{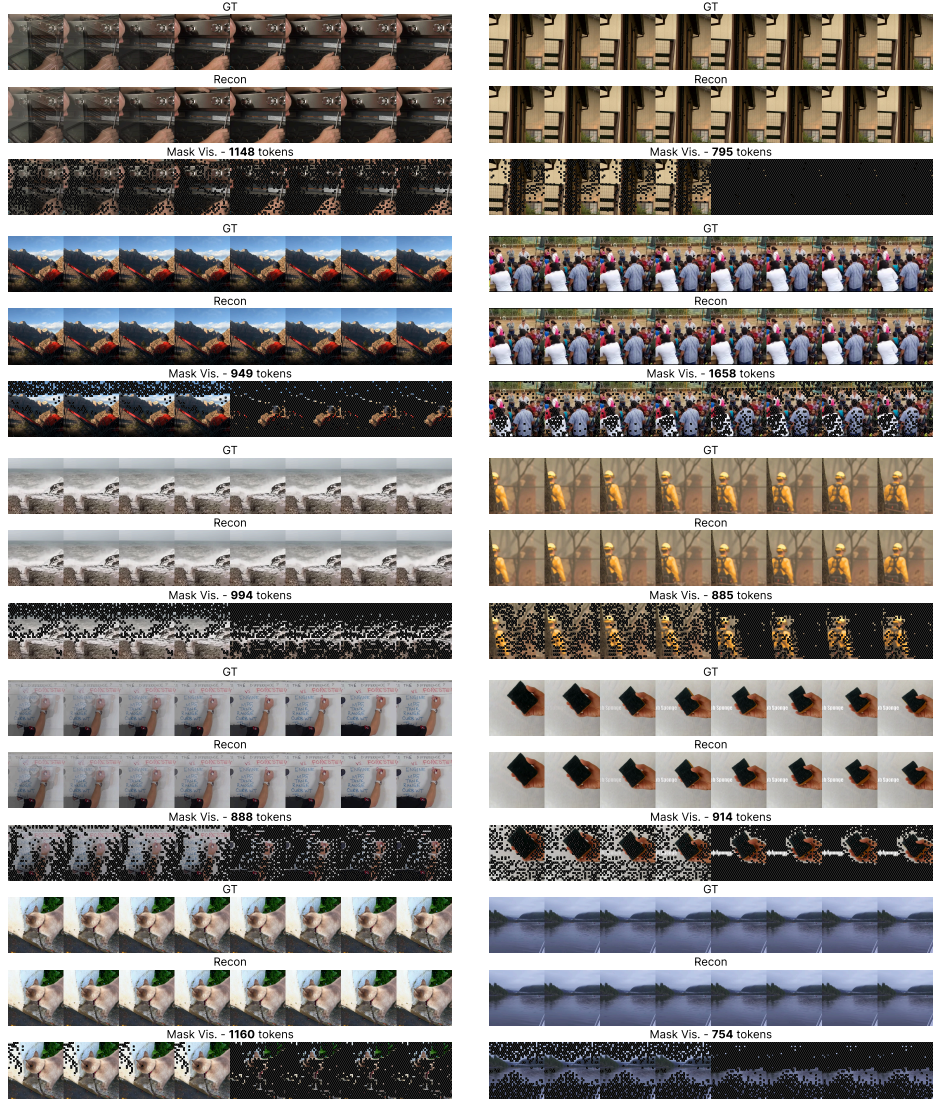}
    \caption{
\textbf{Qualitative reconstruction results.}
Examples are sampled from the Panda70M validation set. Each sample shows reconstruction of $512^2 \times 16$ videos visualized with a 2-frame skip. For each example, we visualize the ground truth (GT), reconstructed frames (Recon), and the corresponding mask visualization (Mask Vis.) along with the number of tokens used. The maximum token budget is 2050. The results illustrate that the model preserves fine details while adaptively allocating tokens according to motion and texture complexity.
}
    \label{fig:suppl_recon}
\end{figure}

\paragraph{VAE Reconstruction.} Figure~\ref{fig:suppl_recon} shows additional qualitative reconstruction results, including the ground truth (GT), reconstructed outputs, and corresponding mask visualizations.
The examples demonstrate that the proposed adaptive tokenizer effectively preserves spatio–temporal details while discarding redundant regions across diverse scenes.

\clearpage
\begin{figure}[p]
    \centering
    \includegraphics[width=\textwidth]{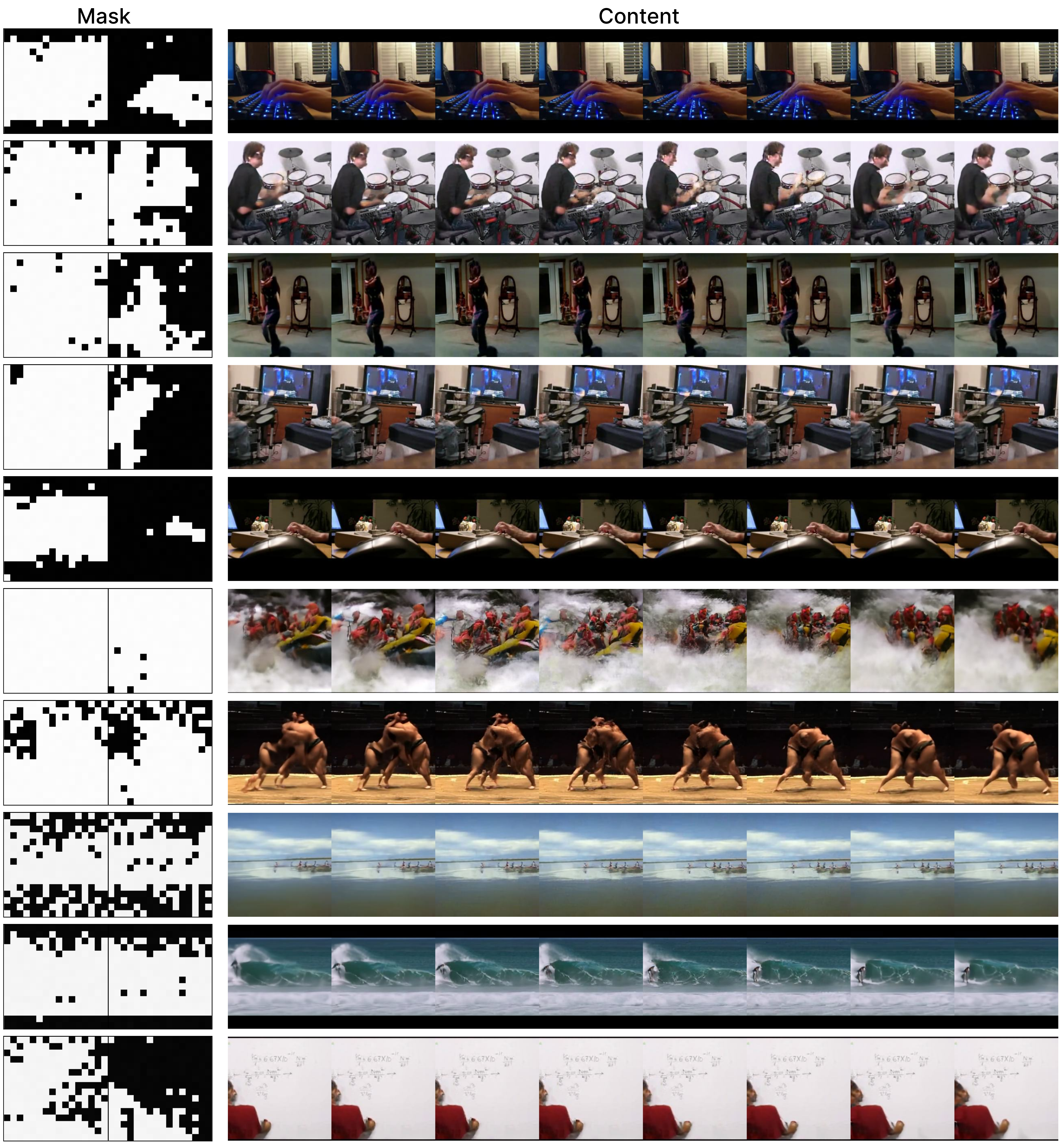}
    \caption{
\textbf{Qualitative video generation results with mask visualization on UCF-101.}
Each row shows a sample generated by the cascaded generation pipeline at $256^2 \times 16$ resolution, visualized with a 2-frame skip.
The left column displays the token selection mask generated by the mask prior (white = active, black = dropped), and the right column shows the corresponding generated video frames.
The cascaded model adaptively allocates more tokens to regions with complex motion and fine-grained details, while suppressing tokens in static or homogeneous areas.
}
    \label{fig:suppl_gen_w_mask_ucf}
\end{figure}

\begin{figure}[p]
    \centering
    \includegraphics[width=\textwidth]{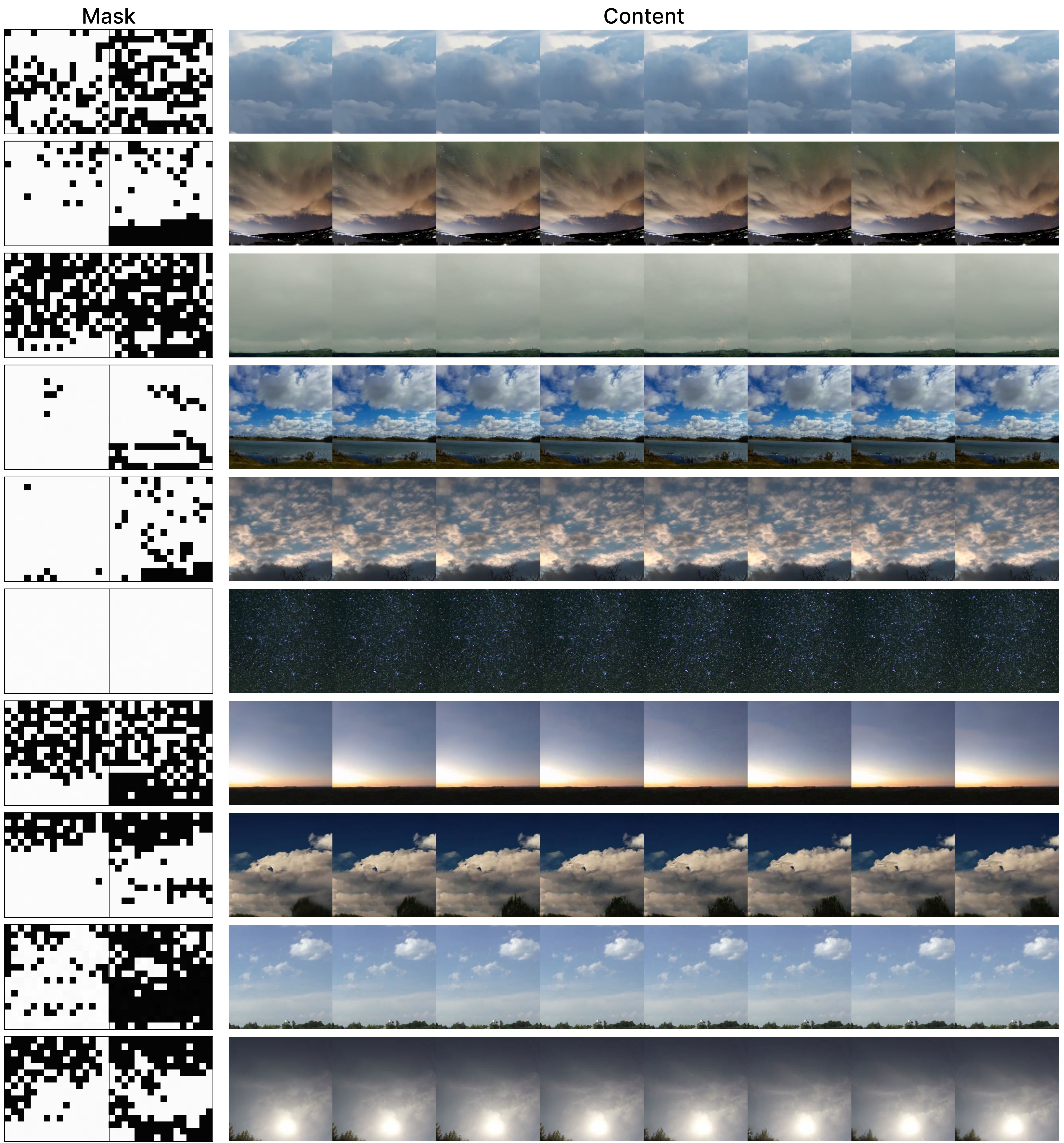}
    \caption{
\textbf{Qualitative video generation results with mask visualization on SkyTimelapse.}
Same setup as Fig.~\ref{fig:suppl_gen_w_mask_ucf}.
Compared to UCF-101, the generated masks tend to be sparser, reflecting the relatively static nature of sky scenes.
The model uses fewer tokens for uniform regions (e.g., clear skies) and allocates more tokens where cloud textures or horizon details require finer representation.
}
    \label{fig:suppl_gen_w_mask_sky}
\end{figure}

\begin{figure}[p]
    \centering
    \includegraphics[width=\textwidth]{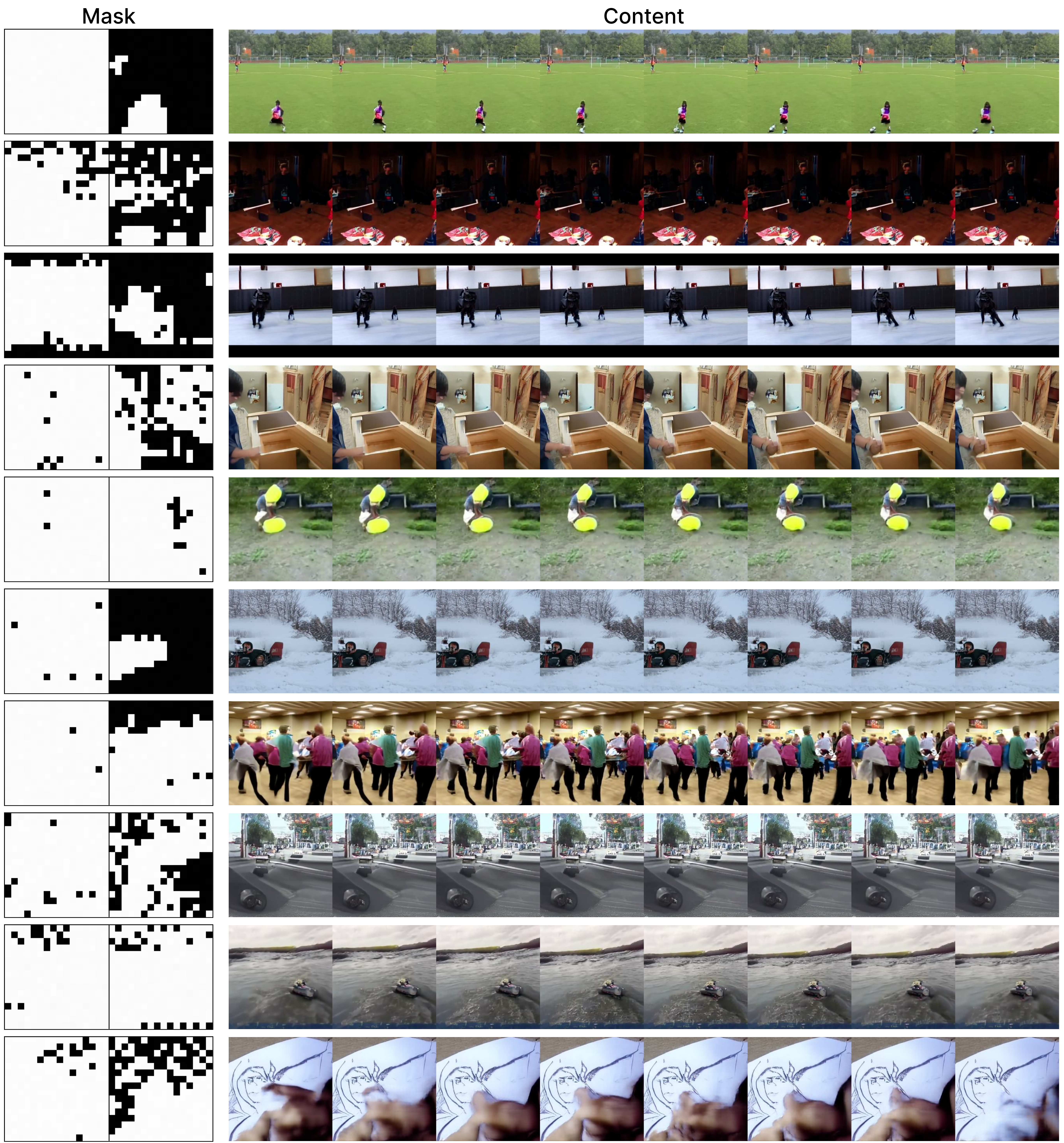}
    \caption{
\textbf{Qualitative video generation results with mask visualization on Kinetics-600.}
Same setup as Fig.~\ref{fig:suppl_gen_w_mask_ucf}.
Kinetics-600 contains diverse real-world activities, resulting in highly varied mask patterns across samples.
Scenes with large camera or object motion retain more tokens, while static backgrounds are aggressively pruned.
}
    \label{fig:suppl_gen_w_mask_kin}
\end{figure}
\clearpage

\begin{figure}[!thbp]
    \centering
    \includegraphics[width=\textwidth]{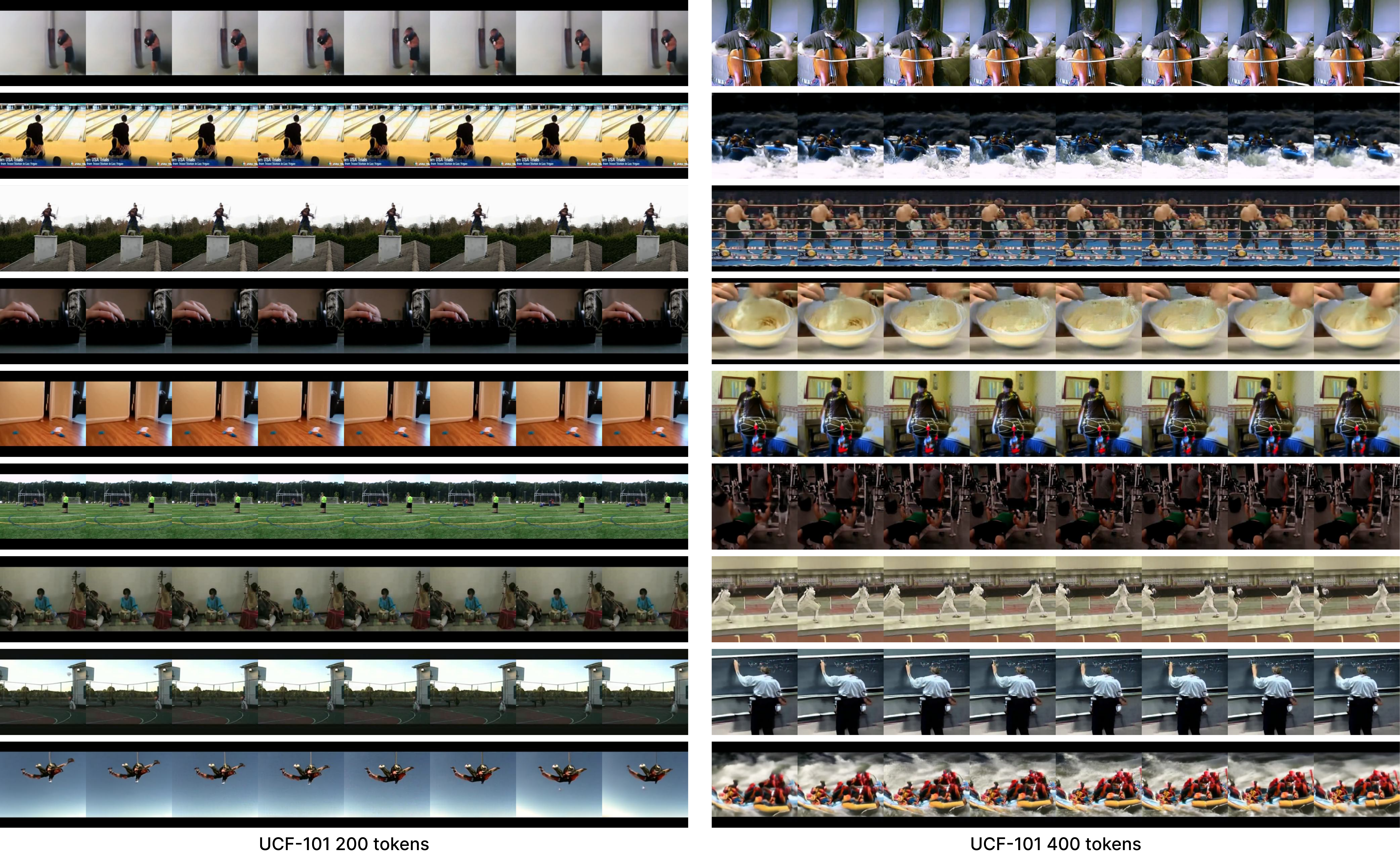}
    \caption{\textbf{Effect of token budget on video generation (UCF-101).}
Qualitative comparison of videos generated with 200 tokens (left) and 400 tokens (right) on the UCF-101 dataset. All samples are generated at $256^2 \times 16$ resolution and visualized with a 2-frame skip. With fewer tokens, the model generates simpler scenes with less motion, while increasing the token budget yields richer motion dynamics and finer visual details.
}
    \label{fig:suppl_gen_tokens_ucf}
\end{figure}

\begin{figure}[!bhtp]
    \centering
    \includegraphics[width=\textwidth]{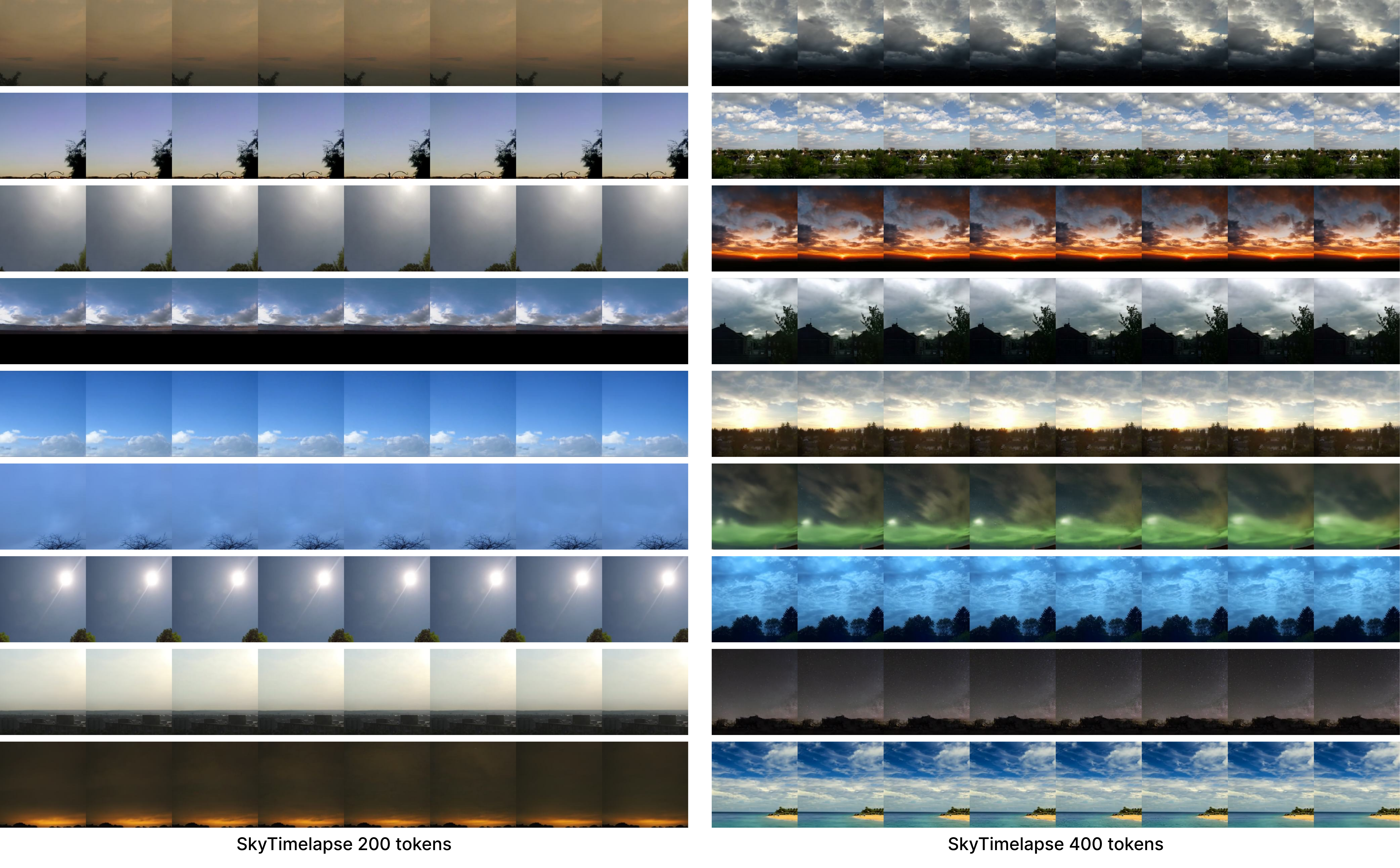}
    \caption{\textbf{Effect of token budget on video generation (SkyTimelapse).}
Qualitative comparison of videos generated with 200 tokens (left) and 400 tokens (right) on the SkyTimelapse dataset. All samples are generated at $256^2 \times 16$ resolution and visualized with a 2-frame skip. With more tokens, the model produces more complex cloud formations and richer atmospheric variations.
}
    \label{fig:suppl_gen_tokens_sky}
\end{figure}

\begin{figure}[!htbp]
    \centering
    \includegraphics[width=\textwidth]{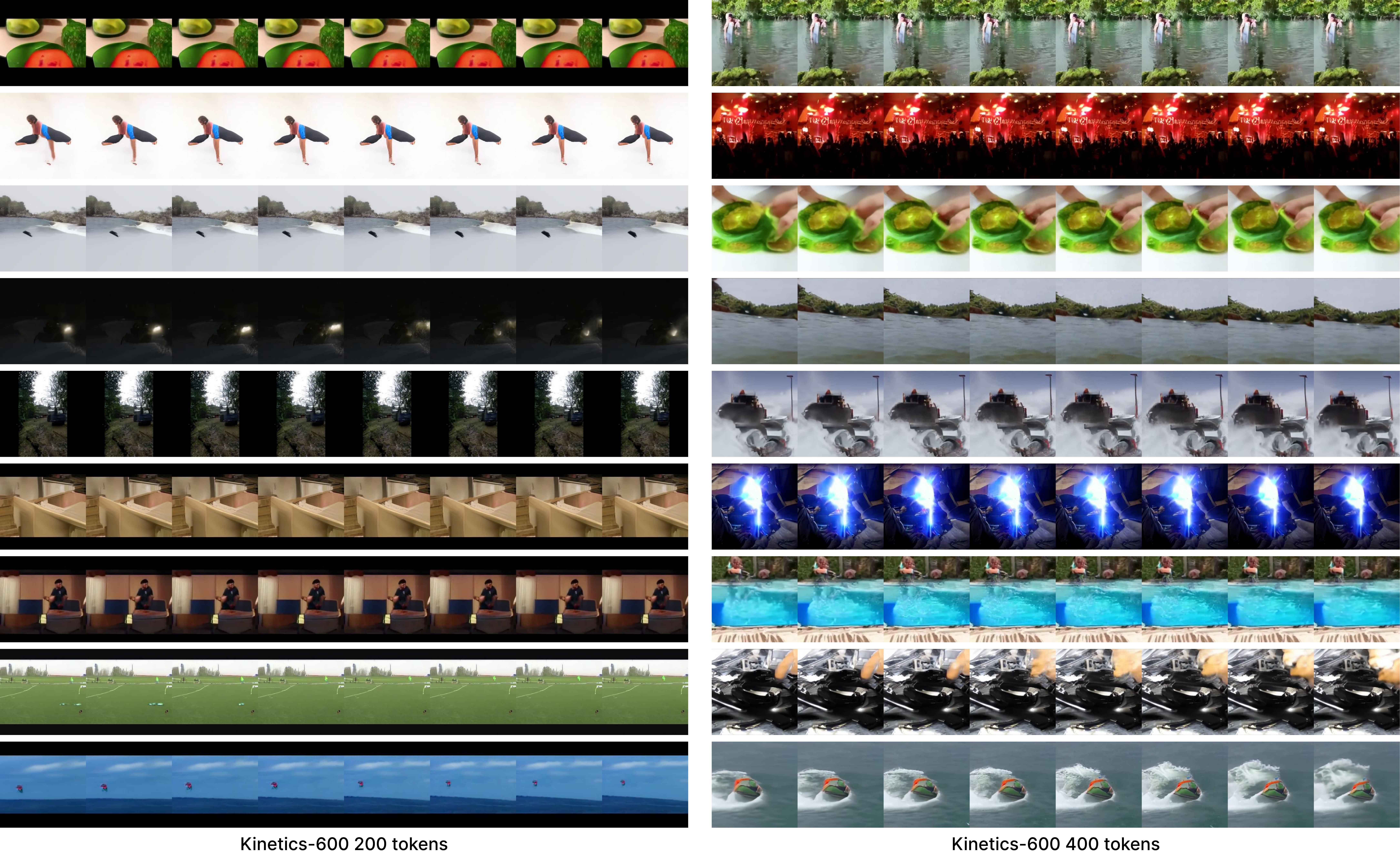}
    \caption{\textbf{Effect of token budget on video generation (Kinetics-600).}
Qualitative comparison of videos generated with 200 tokens (left) and 400 tokens (right) on the Kinetics-600 dataset. All samples are generated at $256^2 \times 16$ resolution and visualized with a 2-frame skip. Higher token budgets enable the model to capture more complex activities and scene details characteristic of this diverse dataset.
}
    \label{fig:suppl_gen_tokens_kin}
\end{figure}

\paragraph{Video Generation with Mask Visualization.} Figures~\ref{fig:suppl_gen_w_mask_ucf}, \ref{fig:suppl_gen_w_mask_sky}, and~\ref{fig:suppl_gen_w_mask_kin} present qualitative generation results produced by the cascaded generation pipeline on the UCF-101~\cite{soomro2012ucf101}, SkyTimelapse~\cite{dtvnetsky}, and Kinetics-600 \cite{kay2017kinetics} datasets, respectively.
Each sample shows the generated token selection mask alongside the corresponding video frames.
The model produces temporally coherent and visually consistent sequences across a wide range of scenes, with the cascaded model adaptively allocating tokens according to scene complexity.

\paragraph{Effect of Token Budget.}
Figures~\ref{fig:suppl_gen_tokens_ucf}, \ref{fig:suppl_gen_tokens_sky}, and~\ref{fig:suppl_gen_tokens_kin} compare samples generated with different token counts (200 vs.\ 400) on UCF-101, SkyTimelapse, and Kinetics-600, respectively.
Increasing the number of tokens results in richer motion and finer details, while fewer tokens yield simpler, more static generations.

\section{Extended Analyses and Visualizations}

\paragraph{Mask Prior Analysis.}
We analyze the cascaded mask prior along three axes: efficiency, mask quality, and
robustness to mask errors.
\emph{(i) Efficiency.} The mask prior is lightweight, adding only $8.3$M parameters
($1.2\%$ of the $686$M content model) and $4.72\%$ to the total sampling cost, so the
two-stage cascaded pipeline remains nearly as efficient as single-stage generation.
\emph{(ii) Mask quality.} To assess how realistic the predicted masks are, we treat mask
sequences as videos and compute their FVD on two disjoint halves of the Kinetics-600
validation set ($2{,}048$ clips each), keeping one half fixed with ground-truth masks and
varying the other (Table~\ref{tab:suppl_mask_quality}). Predicted masks reach an FVD of
$91.80$, far below the random-mask upper bound ($15{,}405.39$) and even closer to the
ground-truth lower bound ($53.08$) than masks with only $1\%$ of cells flipped ($139.58$),
indicating that the prior captures the structural statistics of real token layouts.
\emph{(iii) Robustness to mask errors.} We further measure how mask errors propagate to the
generated video by flipping a fraction of ground-truth mask cells and reporting the
resulting generation FVD (Table~\ref{tab:suppl_mask_error}). Small perturbations
($1$--$2\%$) cause only minor degradation ($72.63$--$81.26$), and quality degrades
gracefully under larger corruption ($5$--$10\%$), showing that the content model is
tolerant to imperfect mask predictions.

\begin{table}[t]
\centering
\footnotesize
\caption{\textbf{Mask quality measured by FVD on mask sequences.} Mask sequences are
treated as videos; FVD is computed on two disjoint Kinetics-600 validation halves
($2{,}048$ clips each), with one half fixed to ground-truth masks. Lower is better.}
\label{tab:suppl_mask_quality}
\setlength{\tabcolsep}{8pt}
\begin{tabular}{lcccc}
\toprule
Mask source & GT & Predicted & $1\%$-flipped GT & Random \\
\midrule
Mask FVD $\downarrow$ & 53.08 & 91.80 & 139.58 & 15{,}405.39 \\
\bottomrule
\end{tabular}

\vspace{1em}

\caption{\textbf{Error propagation from mask to generation.} Generation FVD when a fraction
of ground-truth mask cells are randomly flipped before content generation.}
\label{tab:suppl_mask_error}
\begin{tabular}{lccccc}
\toprule
Flip ratio & 0\% & 1\% & 2\% & 5\% & 10\% \\
\midrule
gFVD $\downarrow$ & 69.58 & 72.63 & 81.26 & 137.74 & 264.77 \\
\bottomrule
\end{tabular}
\end{table}

\begin{figure}[!htbp]
    \centering
    \includegraphics[width=\textwidth]{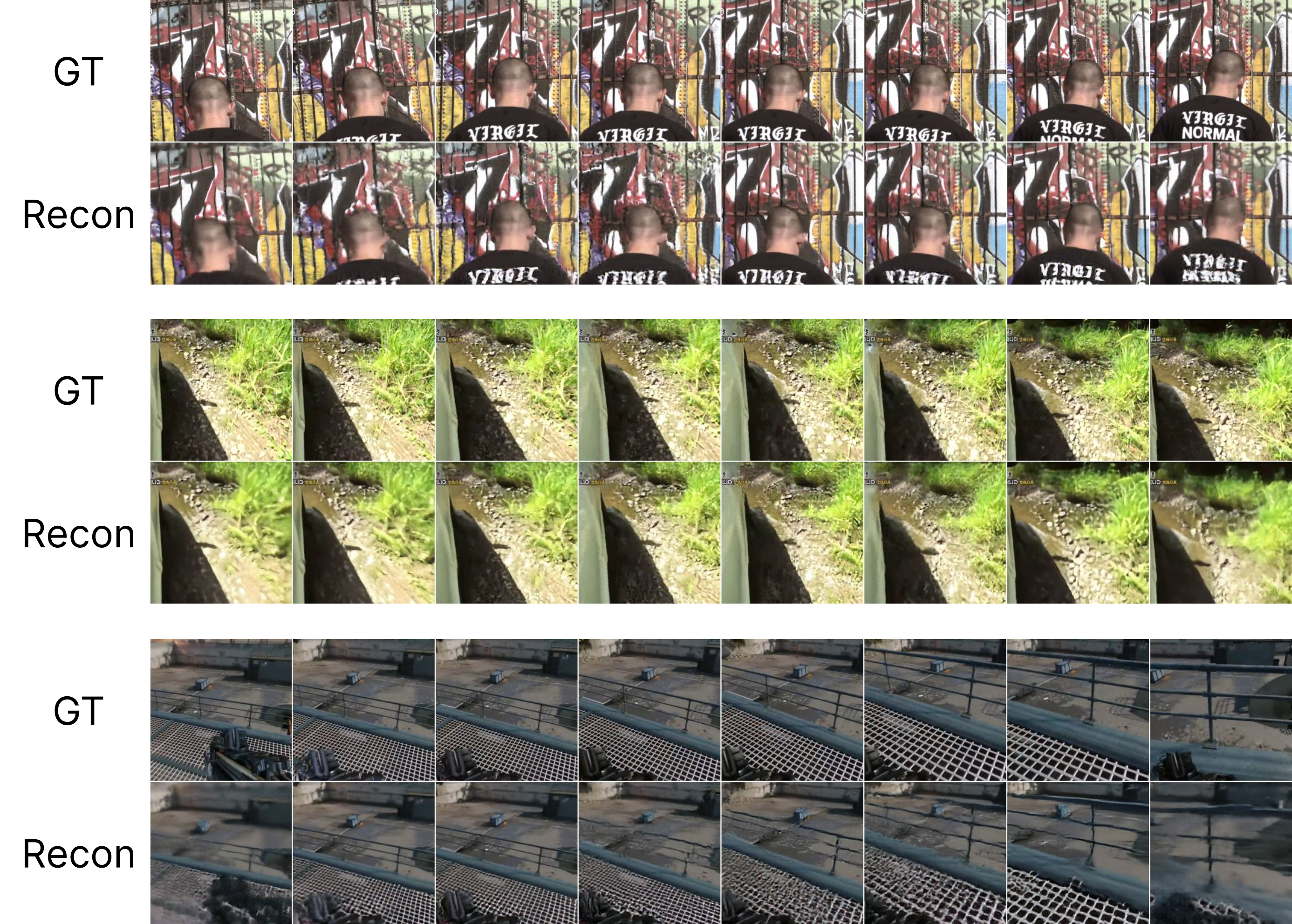}
    \caption{
    \textbf{Failure cases on clips with high-frequency detail and large motion.}
    Each example shows the ground truth (top row, GT) and the reconstruction (bottom row,
    Recon), visualized as $16$ frames sampled with a stride of $2$. On clips that contain
    abundant high-frequency detail together with large motion, the fast-moving fine texture
    is hard to reconstruct faithfully, and the reconstruction becomes blurry precisely in
    the moving high-frequency regions. These samples attain low fidelity
    (PSNR $\approx 19$--$20$) on the Panda-70M validation set.
    }
    \label{fig:suppl_failure}
\end{figure}

\paragraph{Behavior under Domain Shift and Extreme Conditions.}
We examine how KATok behaves outside its training distribution and under extreme inputs.
\emph{(i) Domain shift.} Although the tokenizer is trained only on Panda-70M, its
reconstruction behavior remains consistent on out-of-distribution datasets such as
Kinetics-600 and the egocentric Epic-Kitchens (Fig.~\ref{fig:suppl_cross_recon}), where
token allocation continues to track motion and texture complexity rather than collapsing.
\emph{(ii) Failure cases.} The main failure mode arises on clips that combine abundant
high-frequency detail with large motion. In such regions the fast-moving fine texture is
difficult to reconstruct faithfully, and the reconstruction becomes blurry precisely where
high-frequency content is in motion, yielding low fidelity (PSNR $\approx 19$--$20$ on the
Panda-70M validation set; Fig.~\ref{fig:suppl_failure}).
\emph{(iii) Extreme sparsity.} This regime only arises in generation, where the target token
length is set explicitly; reconstruction infers it adaptively from the input. Requesting an
extremely small budget (fewer than $3$ tokens, well outside the training regime) leaves too
little capacity to encode scene structure, so samples degenerate toward near-uniform,
single-color clips.

\begin{table}[t]
\centering
\footnotesize
\caption{\textbf{Image reconstruction on the \texttt{ImageNet} validation set.}
\#Tokens denotes the average token count.
Comp.$\uparrow$ $= \frac{H \times W \times C}{\text{\#Tokens} \times \text{Channels}}$; not applicable for VQ models (\text{--}).
Ours-Img32 and Ours-Img16 use patch sizes $32^2$ and $16^2$, respectively.
$^\dagger$FlexTok supports variable-length tokenization (1--256) but lacks content-adaptive token selection; we evaluate at its maximum budget of 256 for best performance.}
\label{tab:reconstruction_image}
\setlength{\tabcolsep}{3.5pt}
\begin{tabular}{lccccccc}
\toprule
Method & \#Tokens & Ch. & Comp.$\uparrow$ & PSNR$\uparrow$ & LPIPS$\downarrow$ & SSIM$\uparrow$ & rFID$\downarrow$ \\
\midrule
TiTok-L-VAE   & 32     & 16 & 384    & 19.51 & 0.17 & 0.49 & 1.62 \\
TiTok-B-VAE   & 64     & 16 & 192    & 20.74 & 0.14 & 0.55 & 1.25 \\
TiTok-S-VAE   & 128    & 16 & 96     & 22.51 & 0.09 & 0.64 & 0.84 \\
\midrule
FlexTok d12-d12$^\dagger$ & 256  & -- & --     & 18.47 & 0.22 & 0.47 & 3.18 \\
FlexTok d18-d18$^\dagger$ & 256  & -- & --     & 18.54 & 0.21 & 0.47 & 1.18 \\
FlexTok d18-d28$^\dagger$ & 256  & -- & --     & 18.41 & 0.21 & 0.47 & 1.05 \\
\midrule
ALIT-Small      & 148.15 & 12 & 110.70 & 21.83 & 0.13 & 0.57 & 3.87 \\
Ours-Img32      & 64     & 64 & 48     & 25.38 & 0.12 & 0.73 & 2.97 \\
Ours-Img16      & 229.31 & 32 & 26.83  & \textbf{27.94} & \textbf{0.08} & \textbf{0.83} & \textbf{2.05} \\
\bottomrule
\end{tabular}
\end{table}

\begin{figure}[t]
    \centering
    \includegraphics[width=\columnwidth]{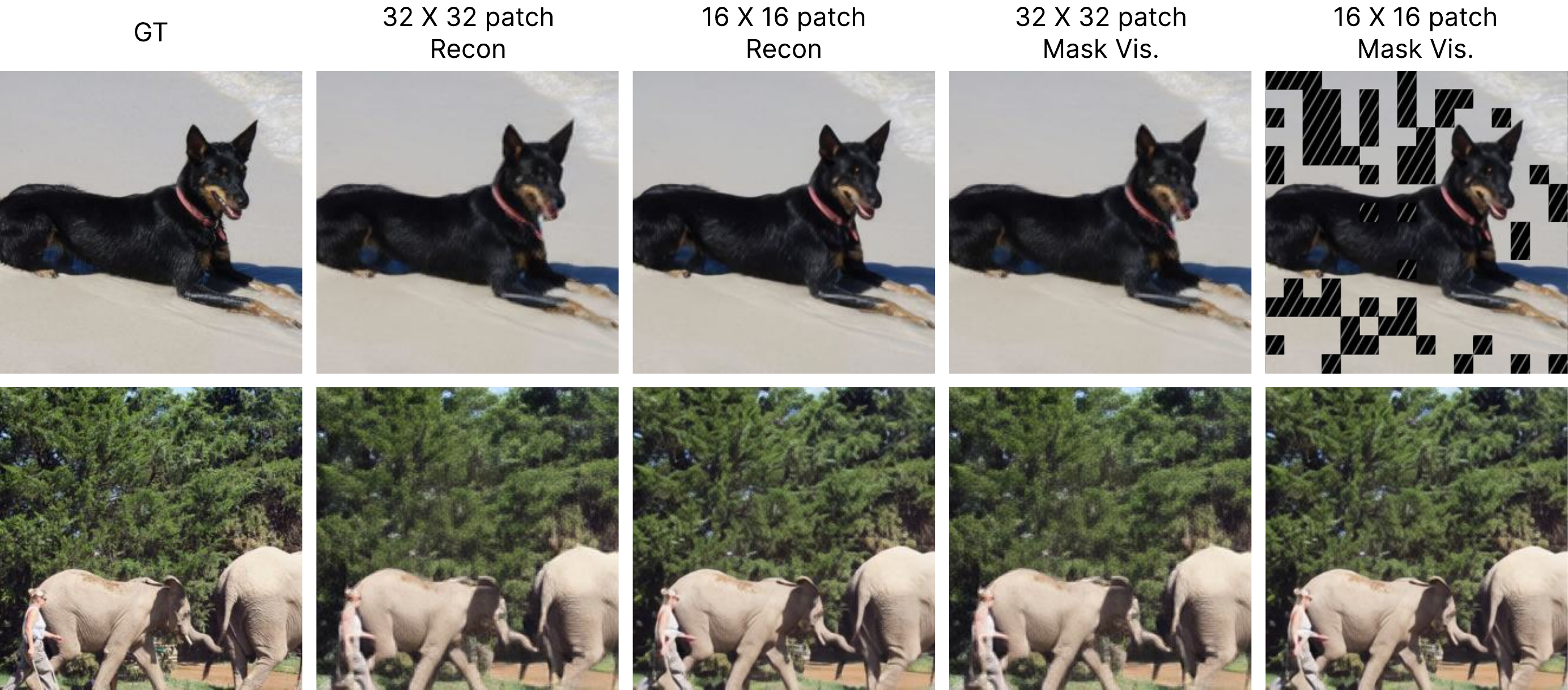}
    \caption{\textbf{ImageNet reconstructions with content-adaptive tokenization.}
Columns show GT, reconstructions with $32{\times}32$ and $16{\times}16$ patches, and the learned token mask for $16{\times}16$ (black tiles = dropped tokens).
Reducing the patch size increases redundancy between neighboring patches, prompting the model to drop more tokens, whereas larger $32{\times}32$ patches retain most tokens.}
    \label{fig:imagenet}
    \vspace{-10pt}
\end{figure}

\paragraph{Image Reconstruction on ImageNet.}
To assess whether the adaptive sparsification mechanism generalizes beyond video, we train image compression models on ImageNet~\cite{deng2009imagenet} using the same architecture and training procedure, varying only the patch size ($32^2$ and $16^2$).
Note that the ImageNet models use symmetric encoder--decoder patch sizes (i.e., no asymmetric coarse-to-fine decoding), as the primary goal here is to isolate the effect of patch-level redundancy on token sparsification rather than to maximize reconstruction quality.
We compare against TiTok~\cite{yu2024titok}, FlexTok~\cite{bachmann2025flextok}, and ALIT~\cite{duggal2025adaptive} using publicly released checkpoints.
For FlexTok, which supports variable-length tokenization but does not provide a mechanism to adapt token counts to input content, we report results at its maximum budget of 256 tokens.

Results are shown in Table~\ref{tab:reconstruction_image}.
The key observation emerges from comparing the two patch sizes.
With $32^2$ patches, each patch covers a large spatial extent and carries substantial unique information, so the model retains nearly all 64 tokens on average.
Reducing the patch size to $16^2$ quadruples the number of initial patches (from 64 to 256), but introduces spatial overlap and local redundancy between neighboring patches.
The adaptive token selector responds accordingly, dropping tokens that encode redundant information and using only 229 out of 256 tokens on average.
This behavior is visible in the mask visualizations of Fig.~\ref{fig:imagenet}: the $16^2$ model drops tokens in homogeneous regions (e.g., uniform backgrounds, smooth textures) while retaining tokens for fine details and object boundaries.

This effect is amplified in the video domain, where an additional temporal dimension of redundancy exists—successive frames share large portions of their content, especially in static or slowly moving scenes.
As a result, our video tokenizer achieves far more aggressive sparsification: on the Panda-70M validation set at $256^2 \times 16$, the model uses an average of 366 tokens out of a maximum of 512, yielding a compression ratio of 134$\times$.
At higher resolutions ($512^2 \times 32$), the compression ratio increases further to 253$\times$, as both spatial and temporal redundancy scale with input size.
In contrast, the image models achieve compression ratios of only 48$\times$ (Ours-Img32) and 27$\times$ (Ours-Img16).
This confirms that our sparsification mechanism directly responds to input redundancy: the richer the redundancy structure of the data modality, the more compact the resulting representation.

\paragraph{Effect of Sparsity Regularization.}

\begin{table}[t]
\centering
\footnotesize
\caption{\textbf{Effect of sparsity regularization weight $\lambda_{\text{sparse}}$.} All models are trained for 70K steps (Stage~1) on Panda-70M with symmetric $16^2 \times 8$ encoder--decoder patches and evaluated on the Panda-70M validation set at $256^2 \times 16$.}
\label{tab:sparsity}
\begin{tabular}{lccccc}
\toprule
$\lambda_{\text{sparse}}$
& 0.005 & 0.010 & 0.020 & 0.050 & 0.100 \\
\midrule
PSNR $\uparrow$  & 28.67 & 28.78 & 22.46 & 21.14 & 19.62 \\
rFVD $\downarrow$ & 50.16 & 50.38 & 462.27 & 680.45 & 863.51 \\
\#Tokens (Avg.) & 401.5 & 372.3 & 34.9 & 16.2 & 6.9 \\
\bottomrule
\end{tabular}
\end{table}

\begin{figure}[!htbp]
    \centering
    \includegraphics[width=\textwidth]{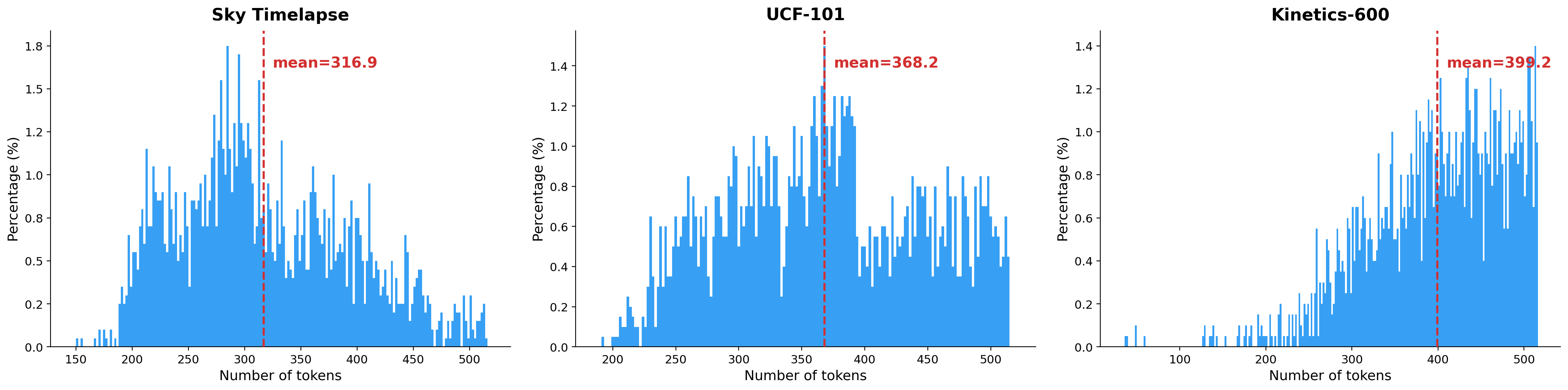}
    \caption{
    \textbf{Token count distributions on SkyTimelapse, UCF-101, and Kinetics-600.}
    Histograms of the number of tokens per video across the entire dataset, with mean token counts indicated by dashed red lines.
    SkyTimelapse uses the fewest tokens on average (316.9), consistent with its predominantly static sky scenes.
    UCF-101 exhibits a higher mean (368.2) due to its greater motion and scene variability.
    Kinetics-600, which contains diverse real-world activities with complex motion, uses the most tokens on average (399.2) and exhibits a wide spread across the full token range with a notable concentration near the maximum budget (512).
    These distributions confirm that the adaptive tokenizer allocates tokens in proportion to content complexity.
    }
    \label{fig:suppl_token_hist}
\end{figure}

Table~\ref{tab:sparsity} reports reconstruction quality and token usage as a function of the sparsity loss weight $\lambda_{\text{sparse}}$, measured after 70K training steps under the coarse decoding setting (i.e., symmetric $16^2 \times 8$ patches for both encoder and decoder, without asymmetric coarse-to-fine decoding).

Mild sparsity ($\lambda_{\text{sparse}} \le 0.01$) achieves a favorable trade-off: the model retains 370--400 tokens on average while maintaining high reconstruction fidelity (PSNR $\ge$ 28.67, rFVD $\le$ 50.38).
A sharp transition occurs between $\lambda_{\text{sparse}} = 0.01$ and $0.02$: the average token count drops from 372 to 35, accompanied by a severe degradation in both PSNR (28.78 $\to$ 22.46) and rFVD (50.38 $\to$ 462.27).
Further increasing the penalty to $\lambda_{\text{sparse}} = 0.1$ reduces token usage to fewer than 7 tokens per clip, but reconstruction quality degrades substantially (PSNR 19.62, rFVD 863.51).
This phase-transition behavior indicates that the model first exploits easy-to-remove redundancy at low penalty, and that beyond a critical threshold the sparsity pressure forces the removal of informationally essential tokens, leading to rapid quality collapse.
Based on this analysis, we adopt $\lambda_{\text{sparse}} = 0.01$ for all main experiments, as it provides the best balance between token efficiency and reconstruction quality.
We did not separately ablate the Gumbel--Softmax temperature $\tau$; the default annealing schedule (from $2.0$ to $0.1$ over the first $10$K steps of Stage~1, without hard discretization) converged stably across all runs, so we kept it fixed for all experiments.

\paragraph{Token Count Distributions.}
Fig.~\ref{fig:suppl_token_hist} shows the token count distributions across all videos in the SkyTimelapse, UCF-101, and Kinetics-600 datasets, tokenized at $256^2 \times 16$ resolution.
The three datasets exhibit clearly distinct distributions that reflect their content characteristics.
SkyTimelapse, which consists primarily of slow-moving cloud and sky scenes, yields the lowest mean token count (316.9 out of 512), with the distribution concentrated in the 200--400 range.
UCF-101 shows a broader and higher distribution (mean 368.2), reflecting the greater diversity of action categories and motion patterns.
Kinetics-600 produces the highest mean token count (399.2) and the widest spread, with a notable concentration near the maximum budget (512), consistent with its large-scale, in-the-wild video collection featuring complex activities, camera motion, and scene transitions.

\begin{figure}[!htbp]
    \centering
    \includegraphics[width=\textwidth]{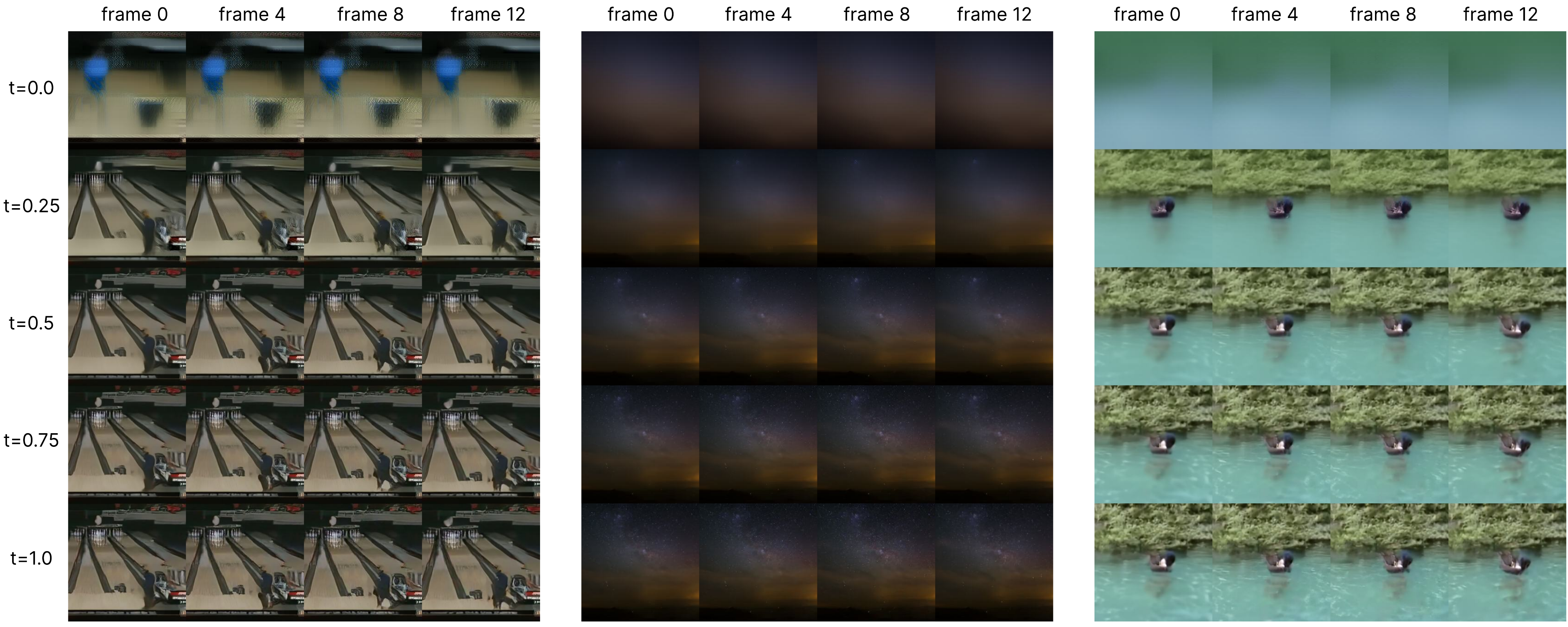}
    \caption{
    \textbf{Step-wise visualization of $x_0$ prediction during generation.}
    Three samples generated on UCF-101 are shown, each visualized at flow-matching timesteps $t \in \{0.0, 0.25, 0.5, 0.75, 1.0\}$ (top to bottom) with frames 0, 4, 8, 12 (left to right).
    At $t=0.0$ (pure noise), the decoded $x_0$ prediction shows only coarse color blobs.
    By $t=0.25$, global scene layout and dominant motion emerge.
    Subsequent steps progressively sharpen spatial details and temporal coherence, with the final output at $t=1.0$ producing a clean, temporally consistent video.
    }
    \label{fig:suppl_x0_pred}
    \vspace{-10pt}
\end{figure}

\paragraph{Denoising Trajectory Visualization.}
Fig.~\ref{fig:suppl_x0_pred} visualizes the intermediate $x_0$ predictions at five flow-matching timesteps ($t = 0.0, 0.25, 0.5, 0.75, 1.0$) for three generated samples on UCF-101.
At $t=0.0$, the model receives pure noise and produces a coarse $x_0$ estimate that captures only low-frequency color distributions.
By $t=0.25$, the global scene structure and dominant motion direction begin to emerge, though fine details remain blurry.
At $t=0.5$, object boundaries, textures, and temporal patterns become recognizable.
The final steps ($t=0.75$ and $t=1.0$) progressively refine high-frequency details and ensure temporal consistency across frames.
This coarse-to-fine generation trajectory demonstrates that the flow-matching model effectively leverages the compact adaptive token representation, first establishing spatial layout and then filling in visual details.

\clearpage
{\small
\bibliographystyle{splncs04}
\bibliography{main} 
}